\documentclass[11pt]{article}
\usepackage[a4paper,total={6.5in,9.5in}]{geometry}

\usepackage[colorlinks=true,linkcolor=blue]{hyperref}

\usepackage{amsmath}
\usepackage{amssymb}
\usepackage{mathtools}
\usepackage{amsthm}
\usepackage{bm}
\usepackage{bbm}
\usepackage{pifont}
\usepackage[dvipsnames]{xcolor}
\newcommand{\cmark}{\ding{51}}
\newcommand{\xmark}{\ding{55}}

\usepackage{microtype}
\usepackage{booktabs}
\usepackage{multirow}

\usepackage{float}
\usepackage{graphicx}
\usepackage{caption}
\usepackage{subcaption}
\graphicspath{{./figures/}}
\DeclareGraphicsExtensions{.pdf}

\usepackage{algorithm}
\usepackage{algorithmic}

\usepackage[capitalize,noabbrev]{cleveref}

\usepackage[numbers]{natbib}

\theoremstyle{plain}
\newtheorem{theorem}{Theorem}[section]
\newtheorem*{theorem*}{Theorem}

\newtheorem{lemma}[theorem]{Lemma}

\theoremstyle{definition}
\newtheorem{definition}[theorem]{Definition}

\theoremstyle{remark}

\def\X{{\mathcal X}}
\def\Y{{\mathcal Y}}
\def\D{{\mathcal D}}

\def\N{{\mathcal N}}
\def\Q{{\mathcal Q}}
\def\Z{{\mathcal Z}}
\def\C{{\mathcal C}}

\def\I{{\mathcal I}}
\def\S{{\mathcal S}}
\def\U{{\mathcal U}}
\def\A{{\mathcal A}}

\def\R{{\mathbb R}}
\def\P{{\mathbb P}}
\def\E{{\mathbb E}}

\def\Id{{\mathbb I}}
\def\1{{\mathbbm 1}}

\def\vlambda{{\bm \lambda}}
\def\veta{{\bm \eta}}
\def\one{{\bm 1}}

\def\tLambda{{\bm{\tilde \Lambda}}}

\def\l{{\hat l}}
\def\u{{\hat u}}
\def\q{{\hat q}}

\def\grad{{\nabla}}
\def\e{{\epsilon}}

\def\d{{\text d}}
\def\cal{{\text{cal}}}
\def\st{{\text{s.t.}}}
\def\opt{{\text{opt}}}
\def\val{{\text{val}}}
\def\rcps{{\text{RCPS}}}
\def\ucb{{\text{UCB}}}

\DeclareMathOperator*{\argmin}{\arg\,\min}

\usepackage[numbers]{natbib}

\title{How to Trust Your Diffusion Model:\\A Convex Optimization Approach to Conformal Risk Control}

\author{
    Jacopo~Teneggi\footnote{Corresponding authors: jtenegg1@jhu.edu , jsulam@jhu.edu .}~\footnote{Department of Computer Science, Johns Hopkins University, Baltimore, MD, 21218}~\footnote{Mathematical Institute for Data Science (MINDS), Johns Hopkins University, Baltimore, MD, 21218} 
    \and Matthew~Tivnan\footnote{Department of Biomedical Engineering, Johns Hopkins University, Baltimore, MD, 21218} 
    \and J. Webster~Stayman\footnotemark[4] 
    \and Jeremias~Sulam~\footnotemark[1]~\footnotemark[4]~\footnotemark[3]
}

\begin{document}
\date{}
\maketitle

\begin{abstract}
    Score-based generative modeling, informally referred to as \emph{diffusion models}, continue to grow in popularity across several important domains and tasks.
    While they provide high-quality and diverse samples from empirical distributions, important questions remain on the reliability and trustworthiness of these sampling procedures for their responsible use in critical scenarios.
    Conformal prediction is a modern tool to construct finite-sample, distribution-free uncertainty guarantees for any black-box predictor.
    In this work, we focus on image-to-image regression tasks and we present a generalization of the Risk-Controlling Prediction Sets (RCPS) procedure, that we term $K$-RCPS, which allows to \emph{(i)} provide entrywise calibrated intervals for future samples of any diffusion model, and \emph{(ii)} control a certain notion of risk with respect to a ground truth image with minimal mean interval length. 
    Differently from existing conformal risk control procedures, ours relies on a novel convex optimization approach that allows for multidimensional risk control while provably minimizing the mean interval length. We illustrate our approach on two real-world image denoising problems: on natural images of faces as well as on computed tomography (CT) scans of the abdomen, demonstrating state of the art performance.
\end{abstract}

\section{Introduction}
Generative modeling is one of the longest standing tasks of classical and modern machine learning \cite{bishop2006pattern}. Recently, the foundational works by \citet{song2019generative,song2020sliced,pang2020efficient} on sampling via \emph{score-matching} \cite{hyvarinen2005estimation} and by \citet{ho2020denoising} on \emph{denoising diffusion models} \cite{sohl2015deep} paved the way for a new class of \emph{score-based} generative models, which solve a reverse-time stochastic differential equation (SDE) \cite{song2020score,anderson1982reverse}. These models have proven remarkably effective on both unconditional (i.e., starting from random noise) and conditional (e.g., inpainting, denoising, super-resolution, or class-conditional) sample generation across a variety of fields \cite{yang2022diffusion,croitoru2022diffusion}. For example, score-based generative models have been applied to inverse problems in general computer vision and medical imaging \cite{kadkhodaie2021stochastic,kawar2021stochastic,kawar2021snips,xie2022measurement,torem2022towards,song2021solving}, 3D shape generation \cite{zeng2022lion,xu2022dream3d,metzer2022latent}, and even in protein design \cite{hoogeboom2022equivariant,corso2022diffdock,watson2022broadly,ingraham2022illuminating}.

These strong empirical results highlight the potential of score-based generative models. However, they currently lack of precise statistical guarantees on the distribution of the generated samples, which hinders their safe deployment in high-stakes scenarios \cite{horwitz2022conffusion}. For example, consider a radiologist who is shown a computed tomography (CT) scan of the abdomen of a patient reconstructed via a score-based generative model. How confident should they be of the fine-grained details of the presented image? Should they trust that the model has not \emph{hallucinated} some of the features (e.g., calcifications, blood vessels, or nodules) involved in the diagnostic process? Put differently, how different will future samples be from the presented image, and how far can we expect them to be from the ground truth image?

In this work we focus on image-to-image regression problems, where we are interested in recovering a high-quality ground truth image given a low-quality observation. While our approach is general, we focus on the problem of image denoising as a running example. We address the questions posed above on the reliability of score-based generative models (and, more generally, of any sampling procedure) through the lens of conformal prediction \cite{papadopoulos2002inductive,vovk2005algorithmic,lei2014distribution,shafer2008tutorial,angelopoulos2021gentle} and conformal risk control \cite{bates2021distribution,angelopoulos2021learn,angelopoulos2022conformal} which provide any black-box predictor with \emph{distribution-free}, \emph{finite-sample} uncertainty guarantees. In particular, the contribution of this paper is three-fold:
\begin{enumerate}
    \item Given a fixed score network, a low-quality observation, and any sampling procedure, we show how to construct valid entrywise calibrated intervals that provide \textit{coverage} of future samples, i.e. future samples (on the same observation) will fall within the intervals with high probability; 
    \item We introduce a novel high-dimensional conformal risk control procedure that minimizes the mean interval length directly, while guaranteeing the number of pixels in the ground truth image that fall outside of these intervals is below a user-specified level on future, unseen low-quality observations;
    \item We showcase our approach for denoising of natural face images as well as for computed tomography of the abdomen, achieving state of the art results in mean interval length.
\end{enumerate}
Going back to our example, providing such uncertainty intervals with provable statistical guarantees would improve the radiologist's trust in the sense that these intervals precisely characterize the type of tissue that could be reconstructed by the model. Lastly, even though our contributions are presented in the context of score-based generative modeling for regression problems---given their recent popularity \cite{kazerouni2022diffusion,yang2022diffusion,croitoru2023diffusion}---our results are broadly applicable to any sampling procedure, and we will comment on potential direct extensions where appropriate.

\subsection{Related work}
\paragraph{Image-to-Image Risk Control} Previous works have explored conformal risk control procedures for image-to-image regression tasks. In particular, \citet{angelopoulos2022image} show how to construct set predictors from heuristic notions of uncertainty (e.g., quantile regression \cite{koenker1978regression,romano2019conformalized}) for any image regressor, and how to calibrate the resulting intervals according to the original RCPS procedure of \citet{bates2021distribution}. \citet{kutiel2022s} move beyond set predictors and propose a mask-based conformal risk control procedure that allows for notions of distance between the ground truth and predicted images other than interval-based ones. Finally, and most closely to this paper, \citet{horwitz2022conffusion} sketch ideas of conformal risk control for diffusion models with the intention to integrate quantile regression and produce heuristic sampling bounds without the need to sample several times. \citet{horwitz2022conffusion} also use the original RCPS procedure to guarantee risk control. Although similar in spirit, the contribution of this paper focuses on a high-dimensional generalization of the original RCPS procedure that formally minimizes the mean interval length. Our proposed procedure is agnostic of the notion of uncertainty chosen to construct the necessary set predictors.

\section{Background}
First, we briefly introduce the necessary notation and general background information. Herein, we will refer to images as vectors in $\R^d$, such that $\X \subset \R^d$ and $\Y \subset \R^d$ indicate the space of high-quality ground truth images, and low-quality observations, respectively. We assume both $\X$ and $\Y$ to be bounded. For a general image-to-image regression problem, given a pair $(x, y)$ drawn from an unknown distribution $\D$ over $\X \times \Y$, the task is to retrieve $x \in \X$ given $y \in \Y$. This is usually carried out by means of a predictor $f:~\Y \to \X$ that minimizes some notion of distance (e.g., MSE loss) between the ground truth images and reconstructed estimates on a set $\{(x_i, y_i)\}_{i=1}^n \sim \D^n$ of $n$ pairs of high- and low-quality images. For example, in the classical denoising problem, one has $y = x + v_0$ where $v_0 \sim \N(0, \sigma_0^2\Id)$ is random Gaussian noise with variance $\sigma_0^2$, and one wishes to learn a denoiser $f$ such that $f(y) \approx x$.

\subsection{\label{sec:conditional_sampling}Score-based Conditional Sampling}
Most image-to-image regression problems are ill-posed: there exist several ground truth images that could have generated the same low-quality observation. This is easy to see for the classical denoising problem described above. Instead of a point predictor $f$---which could approximate a maximum-a-posteriori (MAP) estimate---one is often interested in devising a sampling procedure $F:~\Y \to \X$ for the posterior $p(x | y)$, which precisely describes the distribution of possible ground truth images that generated the observation $y$. In real-world scenarios, however, the full joint $(x, y)$ is unknown, and one must resort to approximate $p(x | y)$ from finite data. It is known that for a general It\^{o} process $\d{x} = h(x, t)\,\d{t} + g(t)\,\d{w}$ that perturbs an input $x$ into random noise \cite{karatzas1991brownian}, it suffices to know the \emph{Stein score} $\grad_x \log p_t(x)$ \cite{anderson1982reverse,liu2016kernelized} to sample from $p(x)$ via the reverse-time process
\begin{equation}
    \label{eq:reverse_time_sde}
    \d{x} = [h(x, t) - g(t)^2 \grad_x \log p_t(x)]\,\d{t} + g(t)\,\d{\bar{w}},
\end{equation}
where $h(x,t)$ and $g(t)$ are a \emph{drift} and \emph{diffusion} term, respectively, and $\d{w}$ and $\d{\bar{w}}$ are forward- and reverse-time standard Brownian motion.\footnote{We will assume time $t$ continuous in $[0, 1]$.} Furthermore, if the likelihood $p(y | x)$ is known---which is usually the case for image-to-image regression problems---it is possible to condition the sampling procedure on an observation $y$. Specifically, by Bayes' rule, it follows that $\grad_x \log p_t(x | y) = \grad_x \log p_t(y | x) + \grad_x \log p_t(x)$ which can be plugged-in into the reverse-time SDE in \cref{eq:reverse_time_sde} to sample from $p(x | y)$. 

Recent advances in generative modeling by \citet{song2019generative,song2020score} showed that one can efficiently train a \emph{time-conditional} score network $s(\tilde{x}, t)$ to approximate the score $\grad_x \log p_t(\tilde{x})$ via \emph{denoising score-matching} \cite{hyvarinen2005estimation}. In this way, given a forward-time SDE that models the observation process, a score network $s(\tilde{x}, t) \approx \grad_x \log p_t(\tilde{x})$, and the likelihood term $p(y | \tilde{x})$, one can sample from $p(x | y)$ by solving the conditional reverse-time SDE with any discretization (e.g., Euler-Maruyama) or predictor-corrector scheme \cite{song2020score}. While these models perform remarkably well in practice, limited guarantees exist on the distributions that they sample from \cite{lee2022convergence}. 
Instead, we will provide guarantees for diffusion models by leveraging ideas of conformal prediction and conformal risk control, which we now introduce.

\subsection{\label{sec:conformal_predictions}Conformal Prediction}
Conformal prediction has a rich history in mathematical statistics \cite{vovk2005algorithmic,papadopoulos2002inductive,vovk2015cross,barber2021predictive,barber2022conformal,gupta2022nested}.\footnote{Throughout this work, we will refer to \emph{split} conformal prediction \cite{vovk2005algorithmic} simply as \emph{conformal prediction}.} It comprises various methodologies to construct finite-sample, statistically valid uncertainty guarantees for general predictors without making any assumption on the distribution of the response (i.e., they are distribution-free). It particular, these methods construct valid prediction sets that provide coverage, which we now define.

\begin{definition}[Coverage \cite{shafer2008tutorial}]
    \label{def:coverage}
    Let $z_1, \dots, z_m, z_{m+1}$ be $m+1$ exchangeable random variables drawn from the same unknown distribution $\Q$ over $\Z$. For a desired miscoverage level $\alpha \in [0, 1]$, a set $\C \subseteq 2^{\Z}$ that only depends on $z_1, \dots, z_m$ provides coverage if
    \begin{equation}
        \P[z_{m+1} \in \C] \geq 1 - \alpha.
    \end{equation}
\end{definition}

We remark that the notion of coverage defined above was introduced in the context of classification problems, where one is interested in guaranteeing that the true, unseen label of a future sample will be in the prediction set $\C$ with high probability. It is immediate to see how conformal prediction conveys a very precise notion of uncertainty---the larger $\C$ has to be in order to guarantee coverage, the more \emph{uncertain} the underlying predictor. We refer the interested reader to \cite{shafer2008tutorial,angelopoulos2021gentle} for classical examples of conformal prediction.

In many scenarios (e.g., regression), the natural notion of uncertainty may be different from miscoverage as described above (e.g., $\ell_2$ norm). We now move onto presenting conformal risk control, which extends the coverage to any notion of risk.

\subsection{\label{sec:conformal_risk_control}Conformal Risk Control}
Let $\I:~\Y \to \X'$  be a general \emph{set-valued} predictor from $\Y$ into $\X' \subseteq 2^{\X}$. Consider a nonnegative loss $\ell:~\X \times \X' \to \R$ measuring the discrepancy between a ground truth $x$ and the predicted intervals $\I(y)$. We might be interested in guaranteeing that this loss will be below a certain tolerance $\e \geq 0$ with high probability on future, unseen samples $y$ for which we do not know the ground truth $x$.
Conformal risk control \cite{bates2021distribution,angelopoulos2021learn,angelopoulos2022conformal} extends ideas of conformal prediction in order to select a specific predictor $\I$ that controls the risk $\E[\ell(x, \I(y))]$ in the following sense.

\begin{definition}[Risk Controlling Prediction Sets]
    \label{def:rcps}
     Let $\S_{\cal} = \{(x_i, y_i)\}_{i=1}^n \sim \D^n$ be a calibration set of $n$ i.i.d. samples from an unknown distribution $\D$ over $\X \times \Y$. For a desired risk level $\e \geq 0$ and a failure probability $\delta \in [0,1]$, a random set-valued predictor $\I:~\Y \to \X' \subseteq 2^{\X}$ is an $(\e, \delta)$-RCPS w.r.t. a loss function $\ell:~\X \times \X' \to \R$ if
    \begin{equation}
        \P_{\S_{\cal}}[\E_{(x,y) \sim \D}[\ell(x, \I(y))] \leq \e] \geq 1 - \delta.
    \end{equation}
\end{definition}

\citet{bates2021distribution} introduced the first conformal risk control procedure for \emph{monotonically nonincreasing} loss functions, those that satisfy, for a fixed $x$,
\begin{equation}
    \label{eq:monotonically_nonincreasing}
    \I(y) \subset \I'(y) \implies \ell(x, \I'(y)) \leq \ell(x, \I(y)).
\end{equation} 
In this way, increasing the size of the sets cannot increase the value of the loss. Furthermore, assume that for a fixed input $y$ the family of set predictors $\{\I_{\lambda}(y)\}_{\lambda \in \Lambda}$, indexed by $\lambda \in \Lambda$, $\Lambda \subset \overline{\R} \coloneqq \R \cup \{\pm \infty\}$, satisfies the following nesting property \cite{gupta2022nested}
\begin{equation}
    \label{eq:nesting}
    \lambda_1 < \lambda_2 \implies \I_{\lambda_1}(y) \subset \I_{\lambda_2}(y).
\end{equation}
Denote $R(\lambda) = \E[\ell(x, \I_{\lambda}(y))]$ the risk of $\I_{\lambda}(y)$ and $\hat{R}(\lambda)$ its empirical estimate over a calibration set $\S_{\cal} = \{(x_i, y_i)\}_{i=1}^n$. Finally, let $\hat{R}^+(\lambda)$ be a \emph{pointwise upper confidence bound} (UCB) that covers the risk, that is
\begin{equation}
    \label{eq:ucb}
    \P[R(\lambda) \leq \hat{R}^+(\lambda)] \geq 1 - \delta
\end{equation}
for \emph{each}, fixed value of $\lambda$---such that can be derived by means of concentration inequalities (e.g., Hoeffding's inequality \cite{hoeffding1994probability}, Bentkus' inequality \cite{bentkus2004hoeffding}, or respective hybridization \cite{bates2021distribution}).\footnote{We stress that \cref{eq:ucb} does \emph{not} imply \emph{uniform coverage} $\forall \lambda \in \Lambda$.} With these elements, \citet{bates2021distribution} show that choosing
\begin{equation}
    \label{eq:rcps}
    \hat{\lambda} = \inf \{\lambda \in \Lambda:~\hat{R}^+(\lambda') < \e,~\forall \lambda' \geq \lambda\}
\end{equation}
guarantees that $\I_{\hat{\lambda}}(y)$ is an $(\e, \delta)$-RCPS according to Definition~\ref{def:rcps}. In other words, choosing $\hat{\lambda}$ as the smallest $\lambda$ such that the UCB is below the desired level $\e$ for all values of $\lambda \geq \hat{\lambda}$ controls the risk al level $\epsilon$ with probability at least $1 - \delta$. For the sake of completeness, we include the original conformal risk control procedure in \cref{algo:rcps} in \cref{supp:rcps}.

Equipped with these general concepts, we now move onto presenting the contributions of this work.

\section{How to Trust Your Diffusion Model}
We now go back to the main focus of this paper: solving image-to-image regression problems with diffusion models. Rather than a point-predictor $f:~\Y \to \X$, we assume to have access to a stochastic sampling procedure $F:~\Y \to \X$ such that $F(y)$ is a random variable with unknown distribution $\Q_y$---that hopefully approximates the posterior distribution of $x$ given $y$, i.e. $\Q_y \approx p(x | y)$. However, we make no assumptions on the quality of this approximation for our results to hold. As described in \cref{sec:conditional_sampling}, $F$ can be obtained by means of a time-conditional score network $s(\tilde{x}, t)$ and a reverse-time SDE. While our results are applicable to \emph{any} sampling procedure, we present them in the context of diffusion models because of their remarkable empirical results and increasing use in critical applications \cite{yang2022diffusion,croitoru2022diffusion}. 

One can identify three separate sources of randomness in a general stochastic image-to-image regression problem: \emph{(i)} the unknown prior $p(x)$ over the space of ground-truth images, as $x \sim p(x)$, \emph{(ii)} the randomness in the observation process of $y$ (which can be modeled by a forward-time SDE over $x$), and finally \emph{(iii)} the stochasticity in the sampling procedure $F(y)$. We will first provide conformal prediction guarantees for a fixed observation $y$, and then move onto conformal risk control for the ground truth image $x$.

\subsection{\label{sec:calibrated_intervals}Calibrated Quantiles for Future Samples}
Given the same low-quality (e.g., noisy) observation $y$, where will future unseen samples from $F(y) \sim \Q_y$ fall? How concentrated will they be? Denote $\I:~\Y \to \X'$ a (random) set-valued predictor from $\Y \subset \R^d$ into a space of sets $\X' \subseteq 2^{\X}$ over $\X \subset \R^d$ (e.g., $\X = [0, 1]^d$, $\X' \subseteq 2^{[0, 1]^d}$). We extend the notion of coverage in Definition~\ref{def:coverage} to \emph{entrywise coverage}, which we now make precise.

\begin{definition}[Entrywise coverage]
    \label{def:entrywise_coverage}
    Let $z_1, \dots, z_m, z_{m+1}$ be $m + 1$ exchangeable random vectors drawn from the same unknown distribution $\Q$ over $\X \subset \R^d$. For a desired miscoverage level $\alpha \in [0, 1]$, a set $\I \subseteq 2^{\X}$ that only depends on $z_1, \dots, z_m$ provides entrywise coverage if
    \begin{equation}
        \P[(z_{m+1})_j \in \I_j] \geq 1 - \alpha
    \end{equation}
    for each $j \in [d] \coloneqq \{1, \dots, d\}$.
\end{definition}

We stress that the definition above is different from notions of \emph{vector quantiles} \cite{carlier2016vector,chernozhukov2017monge} in the sense that coverage is not guaranteed over the entire new random vector $z_{m+1}$ but rather along each dimension independently. Ideas of vector quantile regression (VQR) are complementary to the contribution of the current work and subject of ongoing research \cite{genevay2016stochastic,carlier2020vector,rosenberg2022fast}. 

For a fixed observation $y$, we use conformal prediction to construct a set predictor that provides entrywise coverage.

\begin{lemma}[Calibrated quantiles guarantee entrywise coverage]
    \label{thm:entrywise_coverage}
    Let $F:~\Y \to \X$ be a stochastic sampling procedure from $\Y \subset \R^d$ into $\X \subset \R^d$. Given $y \in \Y$, let $F_1, \dots, F_m, F_{m+1}$ be $m + 1$ i.i.d. samples from $F(y)$. For a desired miscoverage level $\alpha \in [0, 1]$ and for each $j \in [d]$, let $\l_{j,\alpha},\u_{j,\alpha}$ be the $\lfloor (m+1)\alpha/2 \rfloor / m$ and $\lceil (m+1)(1 - \alpha/2) \rceil / m$ entrywise calibrated empirical quantiles of $F_1, \dots, F_m$. Then,
    \begin{equation}
        \label{eq:calibrated_intervals}
        \I^{\alpha}(y)_j = [\l_{j,\alpha}, \u_{j,\alpha}]
    \end{equation}
    provides entrywise coverage.
\end{lemma}

The simple proof of this result is included in \cref{proof:entrywise_coverage}. We remark that, analogously to previous works \cite{angelopoulos2022image,horwitz2022conffusion}, the intervals in $\I^{\alpha}(y)$ are \emph{feature-dependent} and they capture regions of the image where the sampling process $F(y)$ may have larger uncertainty. The intervals in $\I^{\alpha}(y)$ are statistically valid for any number of samples $m$ and any distribution $\Q_y$, i.e. they are not a heuristic notion of uncertainty. If the sampling procedure $F$ is a diffusion model, constructing $\I^{\alpha}(y)$ is agnostic of the discretization scheme used to solve the reverse-time SDE \cite{song2020score} and it does not require retraining the underlying score network, which can be a time-consuming and delicate process, especially when the size of the images is considerable. On the other hand, constructing the intervals $\I^{\alpha}(y)$ requires sampling a large enough number of times from $F(y)$, which may seen cumbersome \cite{horwitz2022conffusion}. This is by construction and intention: diffusion models are indeed very useful in providing good (and varied) samples from the approximate posterior. In this way, practitioners do typically sample several realizations to get an empirical study of this distribution. In these settings, constructing the intervals $\I^{\alpha}(y)$ does not involve any additional computational costs. Furthermore, note that sampling is completely parallelizable, and so no extra complexity is incurred if a larger number of computing nodes are available.

\subsection{\label{sec:convex_conformal_risk_control}A Provable Approach to Optimal Risk Control}
In this section, we will revisit the main ideas around conformal risk control introduced in \cref{sec:conformal_risk_control} and generalize them into our proposed approach, $K$-RCPS. Naturally, one would like a good conformal risk control procedure to yield the shortest possible interval lengths. Assume pixel intensities are normalized between $[0, 1]$ and consider the loss function
\begin{equation}
    \label{eq:01_loss}
    \ell^{01}(x, \I(y)) = \frac{1}{d} \sum_{j \in [d]} \1[x_j \notin \I(y)_j],
\end{equation}
which counts the (average) number of ground truth pixels that fall outside of their respective intervals in $\I(y)$. The constant set-valued predictor $\U(y) = [0, 1]^d$ would trivially control the risk, i.e. $R^{01}(\lambda) = \E[\ell^{01}(x, \U(y))] = 0$. Alas, such a predictor would be completely uninformative. Instead, let $\{\I_{\lambda}(y)\}_{\lambda \in \Lambda}$, $\Lambda \subset \overline{\R}$ be a family of predictors that satisfies the nesting property in \cref{eq:nesting}. In particular, we propose the following additive parametrization in $\lambda$
\begin{equation}
    \label{eq:scalar_additive_interval}
    \I_{\lambda}(y)_j = [\l_j - \lambda, \u_j + \lambda]
\end{equation}
for some lower and upper endpoints $\l_j < \u_j$ that may depend on $y$. For this particularly chosen family of nested predictors, it follows that the mean interval length is
\begin{align}
    \bar{I}(\lambda) = \frac{1}{d} \sum_{j \in [d]}(\u_j - \l_j) + 2\lambda,
\end{align}
a linear function of $\lambda$. Moreover, we can instantiate $\l_j$ and $\u_j$ to be the calibrated quantiles with entrywise coverage, i.e. $\I^{\alpha}_{\lambda}(y) = [\l_{j,\alpha} - \lambda, \u_{j,\alpha} + \lambda]$. 

\begin{figure}[t]
    \centering
    \vspace{-2pt}
    \subcaptionbox{\label{fig:balanced}$\mu = [-1, 1]^T$.}{\includegraphics[width=0.3\linewidth]{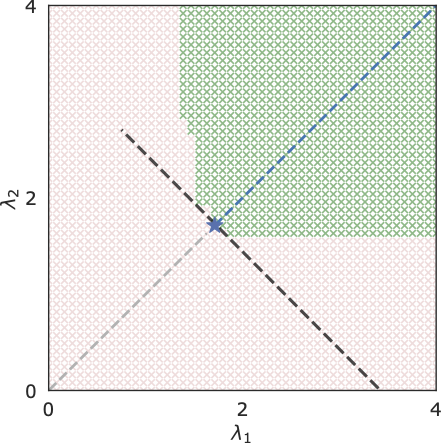}}
    \subcaptionbox{\label{fig:unbalanced}$\mu = [-2, 0.75]^T$.}{\includegraphics[width=0.3\linewidth]{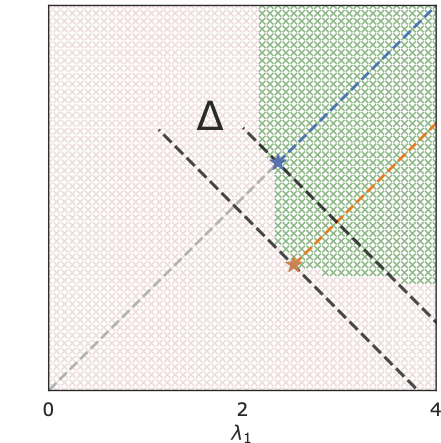}}
    \captionsetup{subrefformat=parens}
    \vspace{-5pt}
    \caption{\label{fig:motivation}Pictorial representation of the suboptimality of the choice of a single scalar parameter $\lambda$ w.r.t. the mean interval length. $\S_{\cal} \sim \N(\mu, \Id_2)^n$, $n = 128$, and $(\I_{\vlambda})_j = [-1 - \lambda_j, 1 + \lambda_j]$, $\vlambda = (\lambda_1, \lambda_2)$. For $\e = \delta = 0.1$, $\hat{R}^{01+}(\vlambda)$ is obtained via Hoeffding-Bentkus hybridization. Green areas indicate regions where $\hat{R}^{01+}(\vlambda) \leq \e$, and conversely for red regions. \subref{fig:balanced} Shows that when features are concentrated symmetrically around the intervals, minimizing $\lambda_1 = \lambda_2 = \lambda$ (blue star) minimizes the mean interval length, while \subref{fig:unbalanced} shows that in the general case, the optimal $\vlambda$ (orange star) may have $\lambda_1 \neq \lambda_2$. $\Delta$ highlights the gain in mean interval length obtained by choosing the orange star instead of the blue one.}
    \vspace{-15pt}
\end{figure}

For such a class of predictors---since the $\ell^{01}$ loss is monotonically nonincreasing---the original RCPS procedure (see \cref{eq:rcps}) is equivalent to the following constrained optimization problem
\begin{align}\tag{P$_1$}
    \label{eq:p01}
    \hat{\lambda}   &= \argmin_{\lambda \in \Lambda}~\bar{I}(\lambda)   \quad\st\quad\hat{R}^{01+}(\lambda') < \epsilon,~\forall \lambda' \geq \lambda 
\end{align}
which naturally minimizes $\lambda$. However, optimizing the mean interval length over a single scalar parameter $\lambda$ is suboptimal in general, as shown in \cref{fig:motivation}. With abuse of notation---we do not generally refer to vectors with boldface---let $\{\I_{\vlambda}(y)\}_{\vlambda \in \Lambda^d}$ be a family of predictors indexed by a $d$-dimensional \emph{vector} $\vlambda = (\lambda_1, \dots, \lambda_d)$ that satisfies the nesting property in \cref{eq:nesting} in an entrywise fashion. A natural extension of \cref{eq:scalar_additive_interval} is then
\begin{equation}
    \label{eq:vector_additive_interval}
    \I_{\vlambda}(y)_j = [\l_j - \lambda_j, \u_j + \lambda_j],
\end{equation}
from which one can define an equivalent function $\bar{I}(\vlambda)$. In particular, using the calibrated intervals as before, define
\begin{equation}
    \I^{\alpha}_{\vlambda}(y) = [\l_{j,\alpha} - \lambda_j, \u_{j,\alpha} + \lambda_j].
\end{equation}
Note now that $\ell^{01}(x, \I_{\vlambda}(y))$ is entrywise monotonically nonincreasing. For fixed vectors $\widetilde{\vlambda} \in \mathbb{R}^d$ and $\veta \in \R^d$ in the positive orthant (i.e., $\veta \geq 0$, entrywise), denote $\tLambda = \widetilde{\vlambda} + \omega\veta,~\omega \in \Lambda$ the set of points on a line with offset $\widetilde{\vlambda}$ and direction $\veta$ parametrized by $\omega$. Then, the $d$-dimensional extension of \eqref{eq:p01} becomes
\begin{align}\tag{P$_d$}
    \label{eq:pd}
    \hat{\vlambda} &= \argmin_{\vlambda \in \tLambda}~\sum_{j \in [d]}\lambda_j    \quad\st\quad\hat{R}^{01+}(\vlambda + \beta\veta) < \epsilon,~\forall \beta \geq 0.
\end{align}
We include an explicit analytical expression for $\hat{R}^{01+}(\vlambda + \veta)$ in \cref{supp:ucb_01}. Intuitively, $\hat{\vlambda}$ minimizes the sum of its entries such that the UCB is smaller than $\e$ for all points \emph{to its right} along the direction of $\veta$ parametrized by $\beta$. We now show  a general high-dimensional risk control result that holds for any entrywise monotonically nonincreasing loss function $\ell$ (and not just $\ell^{01}$ as presented in \eqref{eq:pd}) with risk $R(\vlambda)$, empirical estimate $\hat{R}(\vlambda)$ and respective UCB $\hat{R}^+(\vlambda)$.\footnote{The version of \cref{thm:multidimensional_risk_control} in this manuscript differs from the published paper at \url{https://proceedings.mlr.press/v202/teneggi23a.html}, which has a typo. We detail the subtle differences and implications in \cref{supp:errata}.}

\begin{figure}[t]
    \centering
    \vspace{-2pt}
    \subcaptionbox{\label{fig:loss_x}$\ell^{01},\ell^{\gamma}$ as a function of $x$.}{\includegraphics[width=0.3\linewidth]{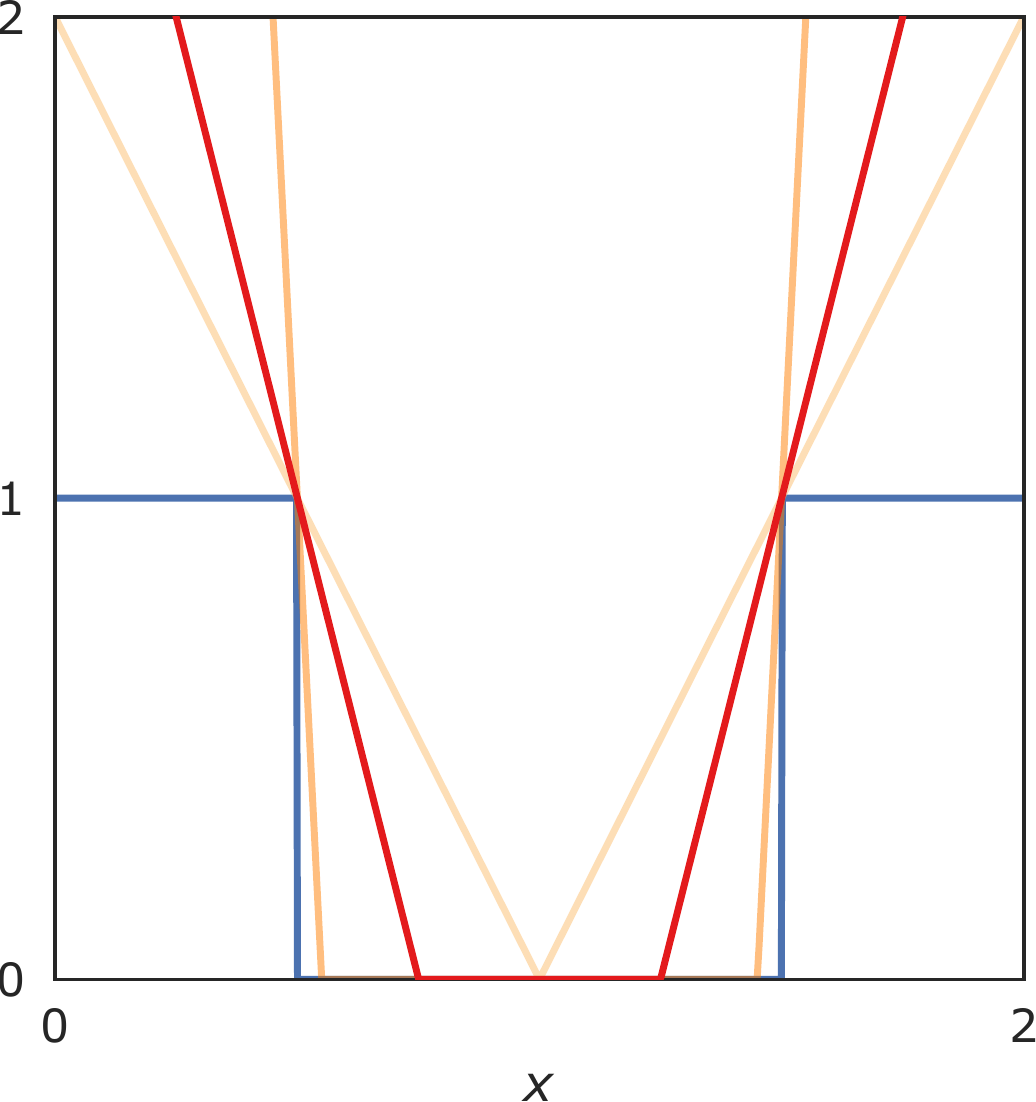}}
    \subcaptionbox{\label{fig:loss_lambda}$\ell^{01},\ell^{\gamma}$ as a function of $\lambda$.}{\includegraphics[width=0.3\linewidth]{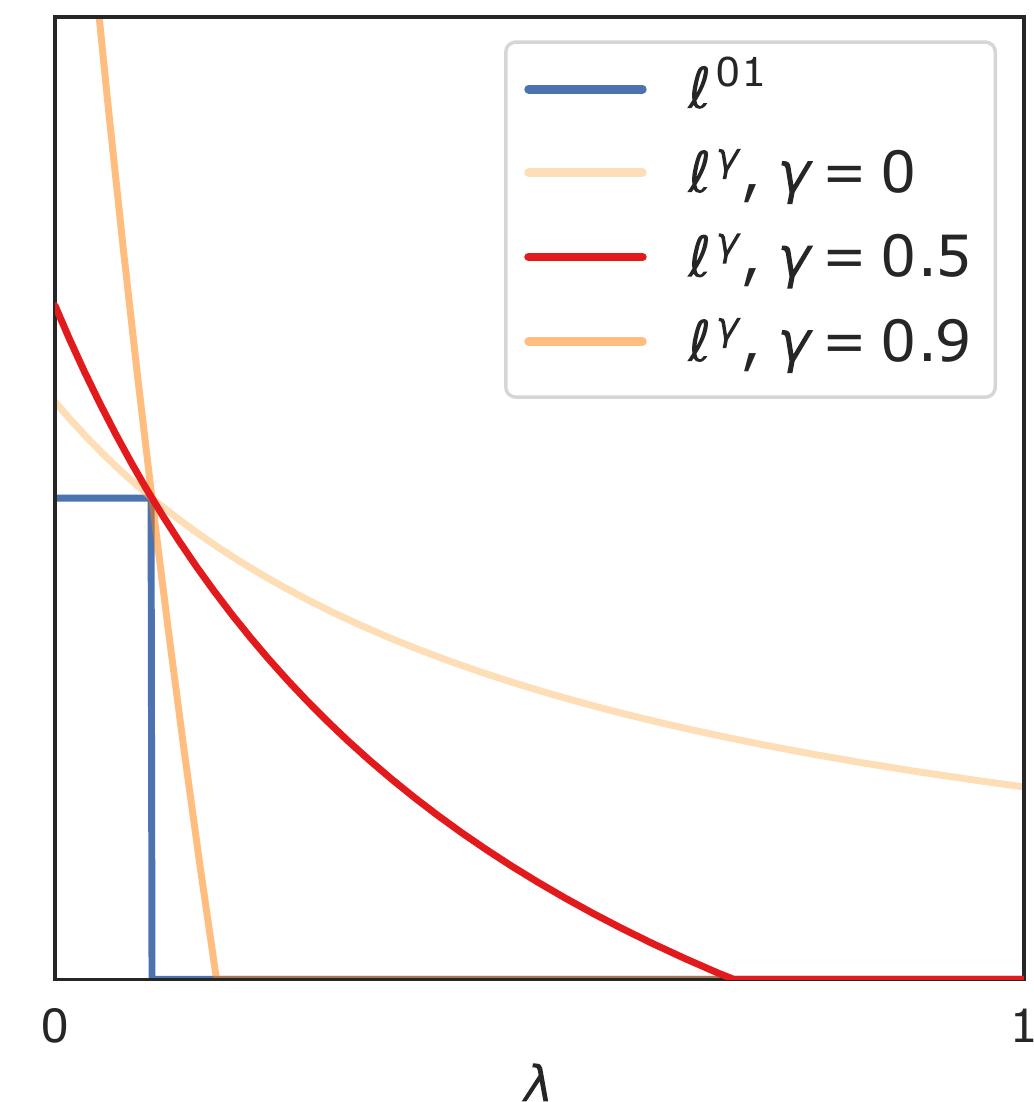}}
    \captionsetup{subrefformat=parens}
    \vspace{-5pt}
    \caption{\label{fig:loss}Visualization of $\ell^{01}(x, \I_{\lambda}(y))$ and $\ell^{\gamma}(x, \I_{\lambda}(y))$ for $\I_{\lambda}(y) = [0.50 - \lambda, 1.50 + \lambda]$, $\gamma \in \{0, 0.5, 0.9\}$. In \subref{fig:loss_x} $\lambda = 0$, and in \subref{fig:loss_lambda} $x = 1.6$.}
    \vspace{-15pt}
\end{figure}

\begin{theorem}[Optimal mean interval length risk control]
    \label{thm:multidimensional_risk_control}
    Let $\ell:~\X \times \X' \to \R$, $\X' \subseteq 2^{\X}$, $\X \subset \R^d$ be an entrywise monotonically nonincreasing function and let $\{\I_{\vlambda}(y) = [\l_j - \lambda_j, \u_j + \lambda_j]\}_{\vlambda \in \Lambda^d}$ be a family of set-valued predictors $\I:~\Y \to \X'$, $\Y \subset \R^d$ indexed by $\vlambda \in \Lambda^d$, $\Lambda \subset \overline{\R}$, for some lower and upper bounds $\l_j < \u_j$ that may depend on $y$. Given $\tLambda = \widetilde{\vlambda} + \omega\veta,~\omega \in \Lambda$, for fixed $\widetilde{\vlambda} \in \R^d$ and $\veta \in \R^d,~\veta \geq 0$, if
    \begin{align}
        \hat{\vlambda}  &= \argmin_{\vlambda \in \tLambda}~\sum_{j \in [d]} \lambda_j   \quad\st\quad\hat{R}^+(\vlambda + \beta\veta) < \e,~\forall \beta \geq 0
    \end{align}
    then $\I_{\hat{\vlambda}}(y)$ is an $(\e, \delta)$-RCPS and $\hat{\vlambda}$ minimizes the mean interval length over $\tLambda$.
\end{theorem}
The proof is included in \cref{proof:multidimensional_risk_control}. Since $\ell^{01}$ is entrywise monotonically nonincreasing, it follows that the solution to \eqref{eq:pd} controls risk. The attentive reader will have noticed (as shown in \cref{fig:motivation}) that the constraint set $\hat{R}^{01+}(\vlambda) \leq \e$ need not be convex. Furthermore, and as shown in \cref{fig:loss_lambda}, $\ell^{01}$ is not convex in $\vlambda$. Hence, it is not possible to optimally solve \eqref{eq:pd} directly. Instead, we relax it to a convex optimization problem by means of a convex upper bound\footnote{We defer proofs to \cref{proof:convexity}.}
\begin{equation}
    \label{eq:gamma_loss}
    \ell^{\gamma}(x, \I_{\vlambda}(y)) = \frac{1}{d} \sum_{j \in [d]} \left[\frac{2(1 + q)}{I(\vlambda)_j} \lvert x_j - c_j \rvert - q\right]_+,
\end{equation}
where $q = \gamma/(1-\gamma)$, $\gamma \in [0, 1)$, $I(\vlambda)_j = \u_j - \l_j + 2\lambda_j$, $c_j = (\u_j + \l_j)/2$, and $[\cdot]_+ = \max(0, \cdot)$. As shown in \cref{fig:loss_x}, the hyperparameter $\gamma$ controls the degree of relaxation by means of changing the portion of the intervals $[\l_j, \u_j]$ where the loss is 0. This way, $\gamma = 0$ retrieves the $\ell_1$ loss centered at $c_j$, and $\lim_{\gamma \to 1} \ell^{\gamma} = \infty$ if $\exists j \in [d]:~x_j \notin [\l_j, \u_j]$ and 0 otherwise.

While one can readily propose a convex alternative to \eqref{eq:pd} by means of this new loss, we instead propose a generalization of this idea in our final problem formulation
\begin{align}\tag{P$_K$}
    \label{eq:pk}
    \widetilde{\vlambda}_K    &= \argmin_{\vlambda \in \Lambda^K}~\sum_{k \in [K]} n_k\lambda_k   \quad\st\quad\hat{R}^{\gamma}(M\vlambda) \leq \e,
\end{align}
for any user-defined $K$-partition of the $[d]$ features---which can be identified by a membership matrix $M \in \{0,1\}^{d \times K}$ where each feature belongs to (only) one of the $K$ groups with $n_k \coloneqq \lvert \{j \in [d]:~M_{jk} = 1\} \rvert$, $\sum_{k \in [K]} n_k = d$. As we will shortly see, it will be useful to define these groups as the empirical quantiles; i.e. set $M$ as that assigning each feature to their respective $k^{\text{th}}$ quantile of the entrywise empirical loss over the optimization set (which is a vector in $\R^d$). We remark that the constrain set in \eqref{eq:pk} is defined on the empirical estimate of the risk of $\I_{\vlambda}(y)$ and it does not involve the computation of the UCB. Then, \eqref{eq:pk} can be solved with any standard off-the-shelf convex optimization software (e.g., CVXPY \cite{diamond2016cvxpy,agrawal2018rewriting}, MOSEK \cite{mosek}). 

Our novel conformal risk control procedure, $K$-RCPS, finds a vector $\hat{\vlambda}_K$ that approximates a solution to the nonconvex optimization problem \eqref{eq:pd} via a two step procedure: 
\begin{enumerate}
    \item First obtaining the optimal solution $\widetilde{\vlambda}_K$ to a user-defined \eqref{eq:pk} problem, and then
    \item Choosing $\hat{\beta} \in \Lambda$ such that
    \begin{equation*}
        \hat{\beta} = \inf \{\beta \in \Lambda:~\hat{R}^{01+}(M\widetilde{\vlambda}_K + \beta'\one) < \e,~\forall \beta' \geq \beta\}
    \end{equation*}
    and return $\hat{\vlambda}_K = M\widetilde{\vlambda}_K + \hat{\beta}\one$.
\end{enumerate}
Intuitively, the $K$-RCPS algorithm is equivalent to performing the original RCPS procedure along the line $M\widetilde{\vlambda}_K + \beta\one$ parametrized by $\beta$. We remark that---as noted in \cref{thm:multidimensional_risk_control}---any choice of $\veta \geq 0$ provides a valid direction along which to perform the RCPS procedure. Here, we choose $\one$ because it is precisely the gradient of the objective function. Future work entails devising more sophisticated algorithms to approximate the solution of \eqref{eq:pd}.

\begin{algorithm}[t]
   \caption{$K$-RCPS}
   \label{algo:krcps}
\begin{algorithmic}[1]
    \STATE {\bfseries Input:} risk level $\e \geq 0$, failure probability $\delta \in [0, 1]$, calibration set $\S_{\cal} = \{(x_i, y_i)\}_{i=1}^n$ of $n$ i.i.d. samples such that $n = n_{\opt} + n_{\rcps}$, membership function $\mathcal{M}$, family of set-valued predictors $\{\I_{\vlambda}(y) = [\l_j - \lambda_j, u_j + \lambda_j]\}_{\vlambda \in \Lambda^d}$, initial (large) value $\beta_{\max}$, stepsize $\d{\beta} > 0$.
    \STATE Split $\S_{\cal}$ into $\S_{\opt}, \S_{\rcps}$
    \STATE $M \gets \mathcal{M}(\S_{\opt})$
    \STATE $\widetilde{\vlambda}_K \gets \texttt{SOLVE-PK}(\S_{\opt}, M)$
    \STATE $\vlambda \gets M\widetilde{\vlambda}_K + \beta_{\max}\one$
    \STATE $\hat{R}^{01+}(\vlambda) \gets 0$
    \WHILE{$\hat{R}^{01+}(\vlambda) \leq \e$}
        \STATE $\vlambda_{\text{prev}} \gets \vlambda$
        \STATE $\vlambda \gets \vlambda - (\d{\beta})\one$
        \STATE $\vlambda \gets [\vlambda]_+$
        \STATE $\hat{R}^{01}(\vlambda) \gets 1/n_{\rcps} \cdot \sum_{(x_i, y_i) \in \S_{\rcps}} \ell^{01}(x_i, \I_{\vlambda}(y_i))$
        \STATE $\hat{R}^{01+}(\vlambda) \gets \text{UCB}(n_{\rcps}, \delta, \hat{R}^{01}(\vlambda))$
    \ENDWHILE
    \STATE $\hat{\vlambda}_K \gets \vlambda_{\text{prev}}$
    \STATE {\bfseries return} $\hat{\vlambda}_K$
\end{algorithmic}
\end{algorithm}

\cref{algo:krcps} implements the $K$-RCPS procedure for any calibration set $\S_{\cal} = \{(x_i, y_i)\}_{i=1}^n$, any general family of set-valued predictors of the form $\{\I_{\vlambda} = [\l_j - \lambda_j, \u_j + \lambda_j]\}_{\vlambda \in \Lambda^d}$, any membership function $\mathcal{M}:~\{\X \times \Y\}^n \to \{0, 1\}^{d \times K}$, and a general $\ucb(n, \delta, \hat{R}(\vlambda))$ that accepts a number of samples $n$, a failure probability $\delta$, and an empirical risk $\hat{R}(\vlambda)$ and that returns a pointwise upper confidence bound $\hat{R}^+(\vlambda)$ that satisfies \cref{eq:ucb}. We remark that, following the \emph{split fixed sequence testing} framework introduced in \citet{angelopoulos2021learn} and applied in previous work \cite{laufer2022efficiently}, the membership matrix and its optimization problem \eqref{eq:pk} are computed on a subset $\S_{\opt}$ of the calibration set $\S_{\cal}$, such that the direction $M \widetilde{\vlambda}_K + \beta\one$ along which to perform the RCPS procedure is chosen \emph{before} looking at the data $\S_{\rcps} = \S_{\cal} \setminus \S_{\opt}$. We note that $K$-RCPS allows for some of the entries in $\hat{\vlambda}_K$ to be set to 0, which preserves the original intervals such that---if they are obtained as described in \cref{sec:calibrated_intervals}---they still provide entrywise coverage of future samples at the desired level $\alpha$.

We now move onto showcasing the advantage of $K$-RCPS in terms on mean interval length on two real-world high dimensional denoising problems: one on natural images of faces as well as on CT scans of the abdomen.

\section{Experiments}

\begin{figure*}[t]
    \centering
    \vspace{-5pt}
    \subcaptionbox{\label{fig:example_calibrated_quantiles_celeba}CelebA dataset, $\sigma_0^2 = 1.0,~\alpha = 0.10$.}{\includegraphics[width=0.48\linewidth]{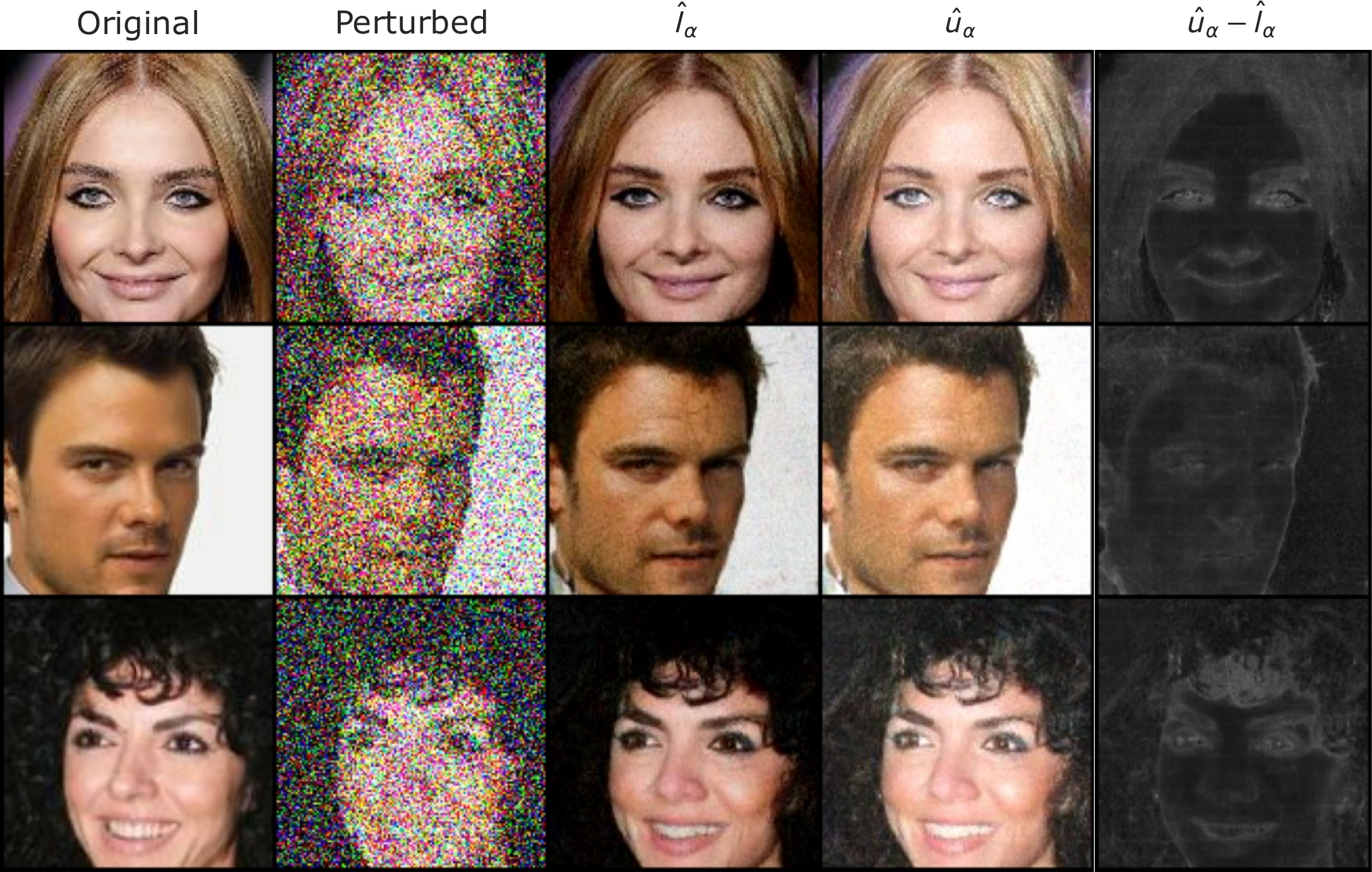}}
    \subcaptionbox{\label{fig:example_calibrated_quantiles_abdomen}AbdomenCT-1K dataset, $\sigma_0^2 = 0.4,~\alpha = 0.20$.}{\includegraphics[width=0.48\linewidth]{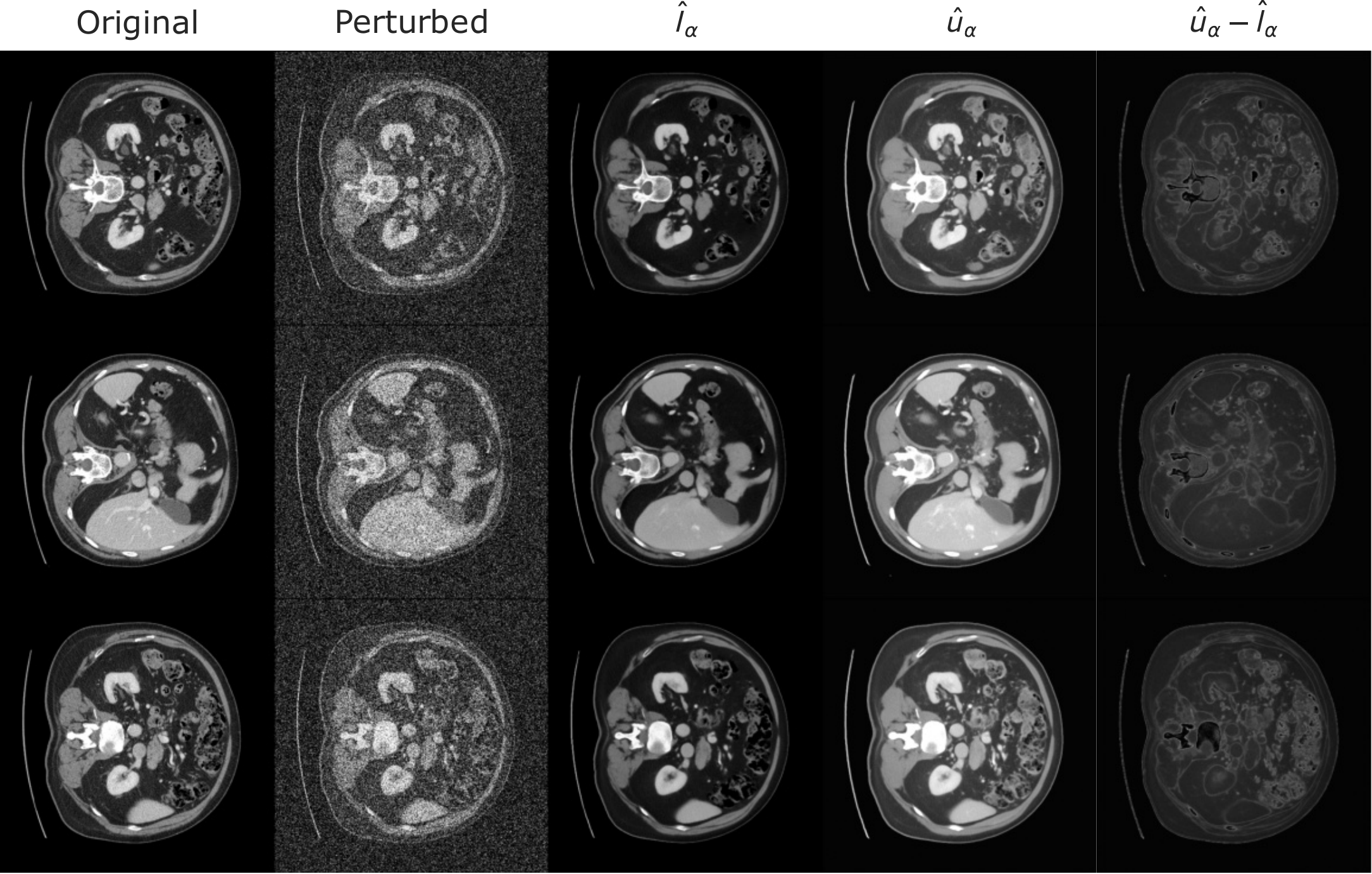}}
    \vspace{-5pt}
    \caption{\label{fig:example_calibrated_quantiles}Calibrated quantiles $\I^{\alpha}(y)$ computed on 128 samples from $F(y)$ for noisy inputs $y$ with noise level $\sigma_0^2$. The difference $\u_{\alpha} - \l_{\alpha}$ represents intervals sizes (i.e., larger intervals indicate larger uncertainty).}
    \vspace{-15pt}
\end{figure*}

As a reminder, the methodological contribution of this paper is two-fold: $(i)$ we propose to use the calibrated quantiles $\I^{\alpha}(y)$ as a statistically valid notion of uncertainty for diffusion models, and $(ii)$ we introduce the $K$-RCPS procedure to guarantee high-dimensional risk control. Although it is natural to use the two in conjunction, we remark that the $K$-RCPS procedure is agnostic of the notion of uncertainty and it can be applied to any nested family of set predictors $\{\I_{\vlambda}(y)\}_{\vlambda \in \Lambda^d}$ that satisfy the additive parametrization in \cref{eq:vector_additive_interval}. Therefore, we compare $K$-RCPS with the original RCPS algorithm on several baseline notions of uncertainty: quantile regression \cite{angelopoulos2022image}, MC-Dropout \cite{gal2016dropout}, N-Con$ffusion$ \cite{horwitz2022conffusion}, and naive (i.e., not calibrated) quantiles. We focus on denoising problems where $y = x + v_0$ with $v_0 \sim \N(0, \sigma_0^2)$, on two imaging datasets: the CelebA dataset \cite{liu2018large} and the AbdomenCT-1K dataset \cite{ma2021abdomenct}. In particular---for each dataset---we train:
\begin{itemize}
    \item A time-conditional score network $s(\tilde{x}, t) \approx \grad_x \log p_t(\tilde{x})$ following \citet{song2020score} to sample from the posterior distribution $p(x | y)$ as described in \cref{sec:conditional_sampling}, and 
    \item a time-conditional image regressor $f:~\Y \times \R \to \X^3$ following \citet{angelopoulos2022image} such that $f(y, t) = (\q_{\alpha/2}, \hat{x}, \q_{1 - \alpha/2})$, where $\hat{x} \approx \E[x \mid y]$ minimizes the MSE loss between the noisy observation $y$ and the ground truth $x$, and $\q_{\alpha/2},\q_{1 - \alpha/2}$ are the $\alpha/2$ and $1 - \alpha/2$ quantile regressors of $x$, respectively \cite{koenker1978regression,romano2019conformalized,angelopoulos2022image}.
\end{itemize}
Both models are composed of the same NCSN++ backbone \cite{song2020score} with dropout $p = 0.10$ for a fair comparison. We then fine-tune the original score network $s(\tilde{x},t)$ according to the N-Con$ffusion$ algorithm proposed by \citet{horwitz2022conffusion} such that---similarly to the image regressor $f$---the resulting time-conditional predictor $\tilde{s}(y, t) = (\q_{\alpha/2}, \q_{1 - \alpha/2})$ estimates the $\alpha/2$ and $1 - \alpha/2$ quantile regressors of $x$. Finally, in order to compare with MC-Dropout, we activate the dropout layers in the image regressor $f$ at inference time, and estimate the mean $\bar{x}$ and standard deviation $\hat{\sigma}$ over 128 samples $\hat{x}_1, \dots, \hat{x}_{128}$. To summarize, we  compare $K$-RCPS and RCPS on the following families of nested set predictors:
\par{\textbf{Quantile Regression (QR)}}
\begin{equation}
    \label{eq:set_qr}
    \I_{\vlambda,\text{QR}}(y)_j = [\hat{x}_j - \lambda_j(\q_{\alpha/2})_j, \hat{x}_j + \lambda_j(\q_{1 - \alpha/2})_j],
\end{equation}
where $f(y, t) = (\q_{\alpha/2}, \hat{x}, \q_{1 - \alpha/2})$.
\par{\textbf{MC-Dropout}}
\begin{equation}
    \label{eq:set_mc_dropout}
    \I_{\vlambda,\text{MC-Dropout}}(y)_j = [\bar{x}_j - \lambda_j\hat{\sigma}_j, \bar{x}_j + \lambda_j\hat{\sigma}_j],
\end{equation}
where $\bar{x},\hat{\sigma}$ are the sample mean and standard deviation over 128 samples $\hat{x}_1, \dots, \hat{x}_{128}$ obtained by activating the dropout layers in the image regressor $f$.
\par{\textbf{N-Con$\bm{ffusion}$}\footnote{A detailed discussion of Con$ffusion$ and the contributions of this paper is included in \cref{supp:comparision_conffusion}.}}
We compare two different parametrizations---multiplicative and additive:
\begin{equation}
    \label{eq:set_conffusion_multiplicative}
    \I^{~\text{multiplicative}}_{\vlambda,\text{Conffusion}}(y)_j = \left[\frac{(\q_{\alpha/2})_j}{\lambda_j}, \lambda_j(\q_{1 - \alpha/2})_j\right]\\
\end{equation}
and
\begin{equation}
    \label{eq:set_conffusion_additive}
    \I^{~\text{additive}}_{\vlambda,\text{Conffusion}}(y)_j = \left[(\q_{\alpha/2})_j - \lambda_j, (\q_{1 - \alpha/2})_j + \lambda_j\right],
\end{equation}
where $\tilde{s}(y, t) = (\q_{\alpha/2}, \q_{1 - \alpha/2})$ is the fine-tuned score network by means of quantile regression on 1000 additional samples.
\par{\textbf{Naive quantiles}}
\begin{equation}
    \label{eq:set_naive_quantiles}
    \I_{\vlambda,\text{naive}}(y)_j = [\l_j - \lambda_j, \u_j + \lambda_j],
\end{equation}
where $\l,\u$ are the naive (i.e., not calibrated) $\alpha/2$ and $1 - \alpha/2$ entrywise empirical quantiles computed on 128 samples from the diffusion model.
\par{\textbf{Calibrated quantiles}}
\begin{equation}
    \label{eq:set_calibrated_quantiles}
    \I^{\alpha}_{\vlambda}(y)_j = [\l_{j,\alpha} - \lambda_j, \u_{j,\alpha} + \lambda_j]
\end{equation}
where $\l_{\alpha},\u_{\alpha}$ are the entrywise calibrated quantiles computed on 128 samples from the diffusion model as described in \cref{eq:calibrated_intervals} (see \cref{fig:example_calibrated_quantiles} for some examples).

\begin{figure*}[t]
    \centering
    \subcaptionbox{\label{fig:example_uncertainty_maps_celeba}CelebA dataset ($\alpha = 0.10$, $\e = 0.10$).}{\includegraphics[width=0.48\linewidth]{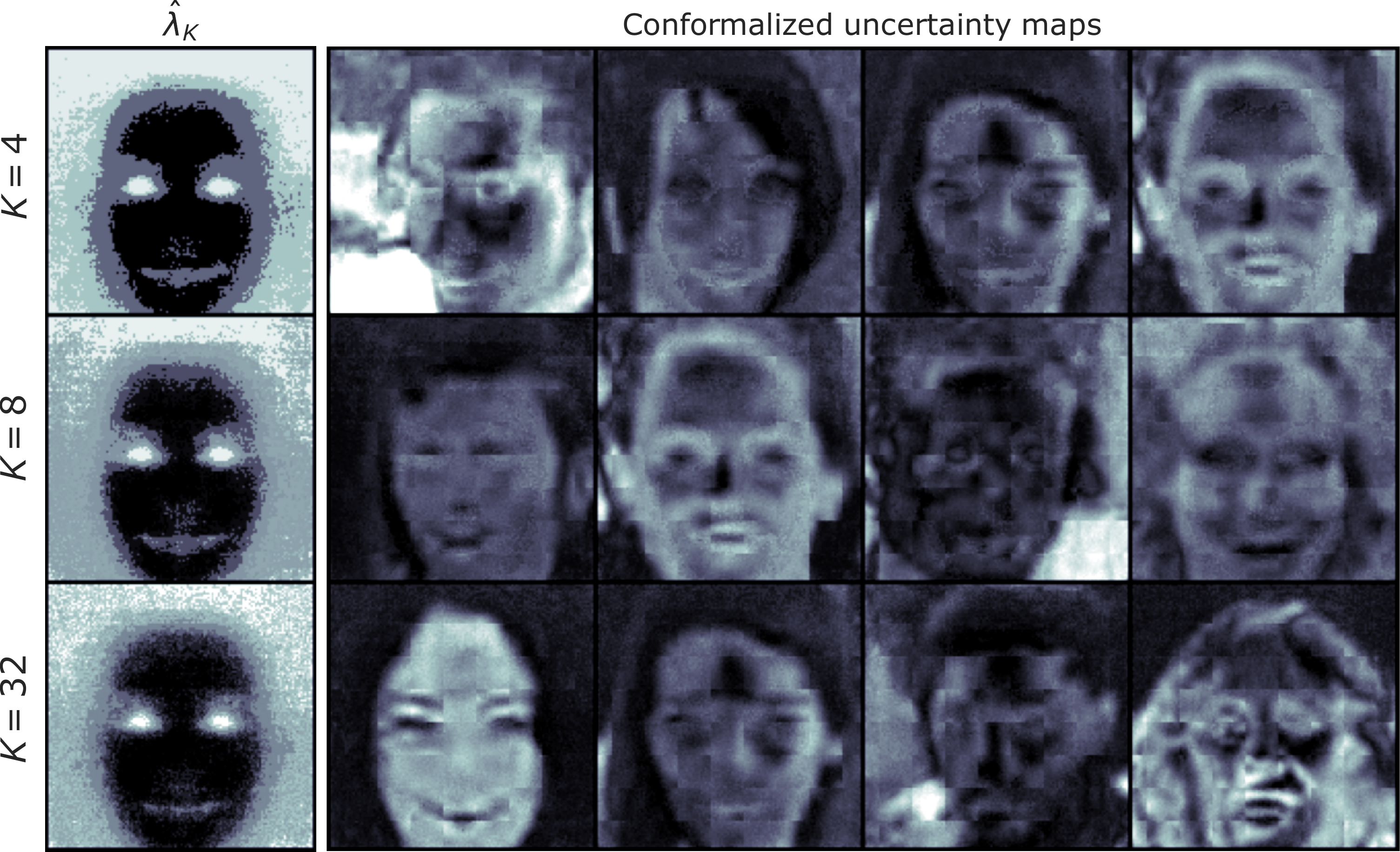}}
    \subcaptionbox{\label{fig:example_uncertinaty_maps_abdomen}AbdomenCT-1K dataset ($\alpha = 0.20$, $\e = 0.05$).}{\includegraphics[width=0.48\linewidth]{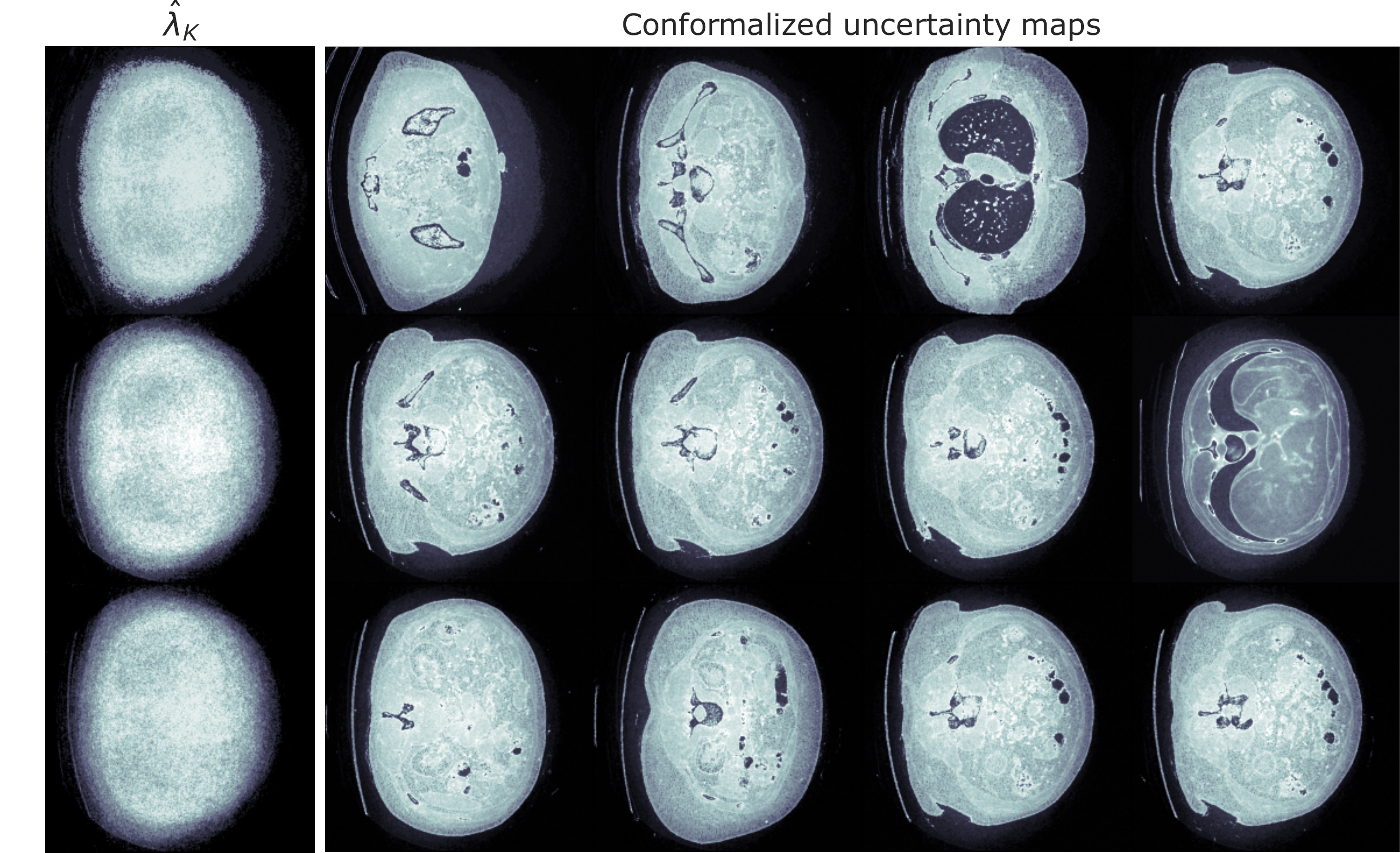}}
    \caption{\label{fig:example_uncertainty_maps}Example optimal $\hat{\vlambda}_K$ for $K \in \{4, 8, 32\}$, $n_{\opt} = 256$, and $d_{\opt} = 100$ with respective conformalized uncertainty maps $\I_{\hat{\vlambda}}^{\alpha}(y) = [\l_{j,\alpha} - (\hat{\lambda}_K)_j, \u_{j,\alpha} + (\hat{\lambda}_K)_j]$. With probability at least $90 \%$ no more than $\e$ portion of the ground truth pixels will fall outside of $\I^{\alpha}_{\hat{\vlambda}}$ on future, unseen samples.}
\end{figure*}

We include further details on the datasets, the models, and the training and sampling procedures in \cref{supp:experimental_details}. {The implementation of $K$-RCPS with all code and data necessary to reproduce the experiments is available at \url{https://github.com/Sulam-Group/k-rcps}.

We compare all models and calibration procedures on 20 random draws of calibration and validation sets $\S_{\cal},\S_{\val}$ of length $n_{\cal}$ and $n_{\val}$, respectively. We remark that for the $K$-RCPS procedure, $n_{\opt}$ samples from $\S_{\cal}$ will be used to solve the optimization problem \eqref{eq:pk}. It follows that for a fixed $n_{\cal}$, the concentration inequality used in the $K$-RCPS procedure will be looser compared to the one in the RCPS algorithm. We will show that there remains a clear benefit of using the $K$-RCPS algorithm in terms of mean interval length given the same amount of calibration data available (i.e., even while the concentration bound becomes looser). In these experiments, we construct the membership matrix $M$ by assigning each feature $j \in [d]$ to the respective $k^{\text{th}}$, $k = 1, \dots, K$ quantile of the entrywise empirical estimate of the risk on $\S_{\opt}$. Furthermore, even though \eqref{eq:pk} is low-dimensional (i.e., $K \ll d$), the number of constraints grows as $d n_{\opt}$, which quickly makes the computation of $\widetilde{\vlambda}_K$ inefficient and time-consuming (e.g., for the AbdomenCT-1K dataset, $d n_{\opt} \sim 10^8$ when $n_{\opt} = 128$, a mild number of samples to optimize over). In practice, we randomly subsample a small number of features $d_{\opt} \ll d$ stratified by membership, which drastically speeds up computation. In the following experiments, we set $d_{\opt} \in \{50, 100\}$ and use $K \in \{4, 8, 32\}$, which reduces the runtime of solving the optimization problem to less than a second for both datasets. Finally, we pick $\gamma$ that minimizes the objective function over 16 values equally spaced in $[0.3, 0.7]$. The choice of these heuristics makes the runtime of $K$-RCPS comparable to that of RCPS, with a small overhead to solve the reduced \eqref{eq:pk} problem (potentially multiple times to optimize $\gamma$).

\cref{fig:risk_control} shows that all combinations of notion of uncertainty and calibration procedures control the risk, as promised. In particular, we set $\delta = 0.10$ for both datasets, and $\e = 0.10, 0.05$ for the CelebA and AbdomenCT-1K dataset, respectively. We repeat all calibration procedures over 20 random samples of $\S_{\cal},\S_{\val}$, with $n_{\val} = 128$, and $n_{\cal} = 640$ or $n_{\cal} = 512$ for the CelebA or AbdomenCT-1K dataset, respectively. \cref{fig:example_uncertainty_maps} showcases some example $\hat{\vlambda}_K$'s obtained by running the $K$-RCPS procedure with $K = 4, 8$, and 32 quantiles alongside their respective conformalized uncertainty maps from $\S_{\val}$. We can appreciate how for both datasets, $\hat{\vlambda}_K$ captures information about the structure of the data distribution (e.g., eyes and lips for the CelebA dataset, and the position of lungs and the heart for the AbdomenCT-1K dataset). Finally, we compare all baselines and calibration procedures in terms of the guarantees each of them provide and their mean interval length. In particular, we report whether each notion of uncertainty provides guarantees over a diffusion model or not. Note that naive and calibrated quantiles are the only notions of uncertainty that precisely provide guarantees on the samples from a diffusion model. Furthermore, calibrated quantiles are the only method that ensures entrywise coverage on future samples on the same noisy observation. For $K$-RCPS, we perform a grid search over $n_{\opt} \in \{ 128, 256\}$, $d_{\opt} \in \{ 50, 100\}$, and $K \in \{ 4, 8, 32\}$, and we report the optimal results in \cref{table:results}. For both datasets, $K$-RCPS provides the tightest intervals among methods that provide both entrywise coverage and risk control for diffusion models. When relaxing the constraint of entrywise coverage, $K$-RCPS still provides the tightest intervals. Across the uncertainty quantification methods that do not relate to a diffusion model, we found that $K$-RCPS with naive sampling provides better results on the CelebA dataset. For the AbdomenCT-1K dataset, N-Con$ffusion$ with multiplicative parametrization and RCPS provides slightly shorter intervals compared to Con$ffusion$ with additive parametrization and $K$-RCPS. However, we stress that the intervals provided by N-Con$ffusion$ are computed for a fine-tuned model that is different from the original diffusion model, and thus provide no guarantees over the samples of the original model. We found that N-Con$ffusion$ with multiplicative parametrization underperforms on the CelebA dataset because the lower bounds do not decay fast enough, and the loss is concentrated on features whose $(\q_{\alpha/2})_j/\lambda_j > x_j$.

\section{Conclusions}

\begin{table*}[t!]
\caption{\label{table:results}Comparison of all notions of uncertainty with RCPS and $K$-RCPS in terms of guarantees provided and mean interval length over 20 independent draws of $\S_{\cal}$. We refer the reader to \cref{supp:comparision_conffusion} for a detailed discussion of the comparison with the Con$ffusion$ framework.}
\centering
\resizebox{\linewidth}{!}{
\begin{tabular}{lcccccc}
\toprule
\multirow{2}{*}{\begin{tabular}[c]{@{}l@{}}Uncertainty\end{tabular}}   & \multirow{2}{*}{\begin{tabular}[c]{@{}c@{}}Diffusion\\ model?\end{tabular}} & \multirow{2}{*}{\begin{tabular}[c]{@{}c@{}}Entrywise\\ coverage?\end{tabular}} & \multirow{2}{*}{\begin{tabular}[c]{@{}c@{}}Risk\\ control?\end{tabular}} & \multirow{2}{*}{\begin{tabular}[c]{@{}c@{}}Calibration\\ procedure\end{tabular}} & \multicolumn{2}{c}{Mean interval length}\\
    &   &   &   &   & CelebA    & AbdomenCT-1K\\
\midrule
QR  & \xmark    & \xmark    & \cmark    & RCPS  & $0.4843 \pm 0.0121$   & $0.2943 \pm 0.0060$\\
MC-Dropout  & \xmark    & \xmark    & \cmark    & RCPS  & $0.6314 \pm 0.0109$ & $0.2810 \pm 0.0013$\\
N-Con$ffusion$  &   &   &   &   &   &\\
--- multiplicative  & \xmark    & \xmark    & \cmark    & RCPS  & $0.6949 \pm 0.0084$   & $0.1126 \pm 0.0020$\\
--- additive    & \xmark    & \xmark    & \cmark    & RCPS  & $0.3314 \pm 0.0040$   & $0.1164 \pm 0.0024$\\
--- additive    & \xmark    & \xmark    & \cmark    & $K$-RCPS  & $0.3131 \pm 0.0056$   & $0.1136 \pm 0.0019$\\
Naive quantiles  & \cmark    & \xmark    & \cmark    & RCPS  & $0.2688 \pm 0.0068$   & $0.1518 \pm 0.0016$\\
Naive quantiles  & \cmark    & \xmark    & \cmark    & $K$-RCPS  & $0.2523 \pm 0.0052$   & $0.1374 \pm 0.0019$\\
\midrule
Calibrated quantiles    & \cmark    & \cmark    & \cmark    & RCPS  & $0.2762 \pm 0.0059$  & $0.1506 \pm 0.0014$\\
Calibrated quantiles    & \cmark    & \cmark    & \cmark    & $K$-RCPS  & $0.2644 \pm 0.0067$ & $0.1369 \pm 0.0016$\\
\bottomrule
\end{tabular}}
\end{table*}

Diffusion models represent huge potential for sampling in inverse problems, alas how to devise precise guarantees on uncertainty has remained open. We have provided \emph{(i)} calibrated intervals that guarantee coverage of future samples generated by diffusion models, \textit{(ii)} shown how to extend RCPS to $K$-RCPS, allowing for greater flexibility by conformalizing in higher dimensions by means of a convex surrogate problem. Yet, our results are general and hold for any data distribution and any sampling procedure---diffusion models or otherwise. When combined, these two contributions provide state of the art uncertainty quantification by controlling risk with minimal mean interval length. Our contributions open the door to a variety of new problems. While we have focused on denoising problems, the application of these tools for other, more challenging restoration tasks is almost direct since no distributional assumptions are employed. The variety of diffusion models for other conditional-sampling problem can readily be applied here too \cite{yang2022diffusion,croitoru2022diffusion}. Lastly---and differently from other works that explore controlling multiple risks \cite{laufer2022efficiently}---ours is the first approach to provide multi-dimensional control of one risk for conformal prediction, and likely improvements to our optimization schemes could be possible. More generally, we envision our tools to contribute to the responsible use of machine learning in modern settings.

\section*{Acknowledgments}
The authors sincerely thank Anastasios Angelopoulos for insightful discussions in early stages of this work, as well as the anonymous reviewers who have helped improve our paper.

\bibliographystyle{plainnat}
\bibliography{bibliography}

\newpage
\appendix
\renewcommand\thefigure{\thesection.\arabic{figure}} 

\section{Proofs}
\setcounter{figure}{0}
In this section, we include the proofs for the results presented in this paper. Herein, denote $\I:~\Y \to \X'$ a set-valued predictor from $\Y \subset \R^d$ into a space of subsets $\X' \subseteq 2^{\X}$ for $\X \subset \R^d$.

\subsection{\label{proof:entrywise_coverage}Proof of Lemma~\ref{thm:entrywise_coverage}}
Let $F:~\Y \to \X$ be a stochastic sampling procedure from $\Y$ to $\X$ such that for a fixed $y \in \Y$, $F(y)$ is a random vector with unknown distribution $\Q_y$. We show that for a desired miscoverage level $\alpha \in [0,1]$, the entrywise calibrated empirical quantiles $\I^{\alpha}(y)_j = [\l_{j,\alpha}, \u_{j,\alpha}]$ defined in \cref{eq:calibrated_intervals} provide entrywise coverage as in Definition~\ref{def:entrywise_coverage}. That is, for each $j \in [d] \coloneqq \{1, \dots, d\}$
\begin{equation}
    \P[F(y)_j \in \I^{\alpha}(y)_j] \geq 1 - \alpha.
\end{equation}
\begin{proof}
    The proof is a variation of the classical split conformal prediction coverage guarantee (see \citet{angelopoulos2021gentle}, Theorem~D.1). Let $F_1, \dots, F_m, F_{m+1}$ be $m+1$ i.i.d. samples from $F(y)$. For a desired miscoverage level $\alpha \in [0, 1]$ and for each $j \in [d]$ denote
    \begin{equation}
        \l_{j,\alpha} = \inf\left\{l:~\frac{\lvert \{k:~(F_k)_j \leq l\} \rvert}{m} \geq \frac{\lfloor (m+1)\alpha/2 \rfloor}{m}\right\}
    \end{equation}
    and
    \begin{equation}
        \u_{j,\alpha} = \inf\left\{u:~\frac{\lvert \{k:~(F_k)_j \leq u\} \rvert}{m} \geq \frac{\lceil (m+1)(1 - \alpha/2) \rceil}{m}\right\}
    \end{equation}
    the $\lfloor (m+1)\alpha/2 \rfloor/m$ and $\lceil (m+1)(1-\alpha/2) \rceil/m$ entrywise calibrated empirical quantiles of $F_1, \dots, F_m$. Assume that for each $j \in [d]$, the first $m$ samples are ordered in ascending order, i.e. $(F_1)_j < \dots < (F_m)_j$ such that
    \begin{equation}
        \l_{j,\alpha} = (F_{\lfloor (m+1)\alpha/2 \rfloor})_j   \quad~\text{and}~\quad  \u_{j,\alpha} = (F_{\lceil (m+1)(1 - \alpha/2) \rceil})_j.
    \end{equation}
    Note that by symmetry of $(F_1)_j, \dots, (F_m)_j$ it follows that $(F_{m+1})_j$ is equally likely to fall between any of the first $m$ samples. That is, for any two indices $m_1 < m_2$
    \begin{equation}
        \P[(F_{m+1})_j \in [(F_{m_1})_j, (F_{m_2})_j]] = \frac{m_2 - m_1}{m+1}.
    \end{equation}
    Instantiating the above equality with $\I^{\alpha}(y)_j = [\l_{j,\alpha}, \u_{j,\alpha}]$ yields
    \begin{align}
        \P[(F_{m+1})_j \in \I^{\alpha}(y)_j]    &= \P\left[(F_{m+1})_j \in [\l_{j,\alpha}, \u_{j,\alpha}]\right]\\
                                                &= \P\left[(F_{m+1})_j \in [(F_{\lfloor (m+1)\alpha/2 \rfloor})_j, (F_{\lceil (m+1)(1 - \alpha/2) \rceil})_j]\right]\\
                                                &= \frac{\lceil (m+1)(1-\alpha/2) \rceil - \lfloor (m+1)\alpha/2 \rfloor}{m+1}\\
                                                &\geq \frac{(m+1)(1-\alpha)}{m+1} = 1 - \alpha
    \end{align}
    which concludes the proof.
\end{proof}

\subsection{\label{proof:multidimensional_risk_control}Proof of \cref{thm:multidimensional_risk_control}}
Recall that for a calibration set $\S_{\cal} = \{(x_i, y_i)\}_{i=1}^n$ of $n$ i.i.d. samples from an unknown distribution $\D$ over $\X \times \Y$, a loss function $\ell:~\X \times \X' \to \R$, and a family $\{\I_{\vlambda}(y)\}_{\vlambda \in \Lambda^d}$ of set-valued predictors indexed by a $d$-dimensional vector $\vlambda = (\lambda_1, \dots, \lambda_d) \in \Lambda^d$, $\Lambda \subset \overline{\R} = \R \cup \{\pm \infty\}$
\begin{equation}
    R(\vlambda) = \E_{(x, y) \sim \D}\left[\ell(x, \I_{\vlambda}(y))\right]  \quad~\text{and}~\quad  \hat{R}(\vlambda) = \frac{1}{n} \sum_{(x_i, y_i) \in \S_{\cal}}^n \ell(x_i, \I_{\vlambda}(y_i))    
\end{equation}
denote the risk of $\I_{\vlambda}(y)$ and its empirical estimate on the calibration set, respectively. Furthermore, let $\hat{R}^+(\vlambda)$ be a pointwise upper confidence bound (UCB) such that for each fixed $\vlambda \in \Lambda^d$ and $\forall \delta \in [0, 1]$
\begin{equation}
    \P[R(\vlambda) \leq \hat{R}^+(\vlambda)] \geq 1-\delta
\end{equation}
as presented in \cref{eq:ucb}. Equivalently to Definition~\ref{def:rcps}, for a risk level $\e \geq 0$, we say that $\I_{\vlambda}(y)$ is an $(\e, \delta)$-RCPS if
\begin{equation}
    \P_{\S_{\cal}}[R(\vlambda) \leq \e] \geq 1 - \delta.
\end{equation}

Given fixed vectors $\widetilde{\vlambda} \in \R^d$ and $\veta \in \R^d,~\veta \geq 0$, denote $\tLambda = \widetilde{\vlambda} + \omega\veta,~\omega \in \Lambda$ the line with offset $\widetilde{\vlambda}$ and direction $\veta$ parametrized by $\omega$. We show that for entrywise monotonically nonincreasing loss functions and for the family of set-valued predictors of the form $\I_{\vlambda}(y) = [\l_j - \lambda_j, \u_j + \lambda_j]$, for some lower and upper bounds $\l_j < \u_j$ that may depend on $y$, if
\begin{equation}
    \label{eq:hat_vlambda}
    \hat{\vlambda} = \argmin_{\vlambda \in \tLambda}~\sum_{j \in [d]} \lambda_j    \quad\st\quad\hat{R}^+(\vlambda + \beta\veta) < \e, \forall \beta \geq 0
\end{equation}
$\I_{\hat{\vlambda}}(y)$ is an $(\e, \delta)$-RCPS. We start by reminding the following definitions.

\begin{definition}[Entrywise monotonically nonincreasing function]
    \label{def:entrywise_monotonically_nonincreasing}
    A loss function $\ell$ is entrywise monotonically nonincreasing if, for a fixed $x$, $\forall j \in [d]$
    \begin{equation}
        \I(y)_j \subset \I'(y)_j \implies \ell(x, \I'(y)) \leq \ell(x, \I(y)).
    \end{equation}
\end{definition}

\begin{definition}[Entrywise nesting property]
    \label{def:entrywise_nesting}
    A family of set predictors $\{\I_{\vlambda}(y)\}_{\vlambda \in \Lambda^d}$ is entrywise nested if, for a fixed $y$, $\forall j \in [d]$
    \begin{equation}
        \lambda_{j,1} < \lambda_{j,2} \implies \I_{[\lambda_{j,1}, \vlambda_{-j}]}(y)_j \subset \I_{[\lambda_{j,2}, \vlambda_{-j}]}(y)_j,
    \end{equation}
    where $[\lambda_j, \vlambda_{-j}]$ is the vector that takes value $\lambda_j$ in its $j^{\text{th}}$ entry and $\lambda_{-j}$ in its complement $-j \coloneqq [d] \setminus \{j\}$.
\end{definition}

\begin{proof}
    The proof is a high-dimensional extension of the validity of the original RCPS calibration procedure. Note that the family of set-valued predictors $\{\I_{\vlambda}(y) = [\l_j - \lambda_j, \u_j + \lambda_j]\}_{\vlambda \in \Lambda^d}$ satisfies the entrywise nesting property in Definition~\ref{def:entrywise_nesting}. For $\hat{\vlambda}$ chosen as in \cref{eq:hat_vlambda}, denote $L:~\Lambda \to \R$ the one-dimensional function such that
    \begin{equation}
        L(\beta) = R(\hat{\vlambda} + \beta\veta)    \quad~\text{and}~\quad  \hat{L}^+(\beta) = \hat{R}^+(\hat{\vlambda} + \beta\veta).
    \end{equation}
    It follows that $L$ is monotonically nonincreasing because $\ell$ is entrywise monotonically nonincreasing by assumption, and $\veta$ belongs to the positive orthant. Furthermore, $\P[L(\beta) \leq \hat{L}^+(\beta)] \geq 1-\delta$ by definition of $\hat{R}^+(\vlambda)$. Denote
    \begin{equation}
        \beta^* = \inf \{\beta \in \Lambda: L(\beta) \leq \e\}
    \end{equation}
    and suppose $R(\hat{\vlambda}) = L(0) > \e$. By monotonicity of $L$ it follows that $\beta^* > 0$, and by definition of $\hat{\vlambda}$, $\hat{L}^+(\beta^*) = \hat{R}^+(\hat{\vlambda} + \beta^*\veta) < \e$. However, since $L(\beta^*) = \e$ and $\P[L(\beta^*) \leq \hat{L}^+(\beta^*)] \geq 1 - \delta$, we conclude that this event happens with probability at most $\delta$. Hence, $\P[R(\hat{\vlambda}) \leq \e] \geq 1 - \delta$ and $\I_{\hat{\vlambda}}(y)$ is an $(\e, \delta)$-RCPS.

    Lastly, it is easy to see that $\hat{\vlambda}$ minimizes the mean interval length $\bar{I}(\vlambda)$ over $\tLambda$. Note that
    \begin{equation}
        \bar{I}(\vlambda) = \frac{1}{d} \sum_{j \in [d]} (\u_j - \l_j) + 2 \sum_{j \in [d]} \lambda_j,
    \end{equation}
    and the statement follows by definition.
\end{proof}

\subsection{\label{proof:convexity}Proof that $\ell^{\gamma}$ is a convex upper bound to $\ell^{01}$ (see \cref{eq:gamma_loss})}
Recall that for a family of set-valued predictors $\{\I_{\vlambda}(y) = [\l_{j} - \lambda_j, \u_j + \lambda_j]\}_{\vlambda \in \Lambda^d}$ indexed by a $d$-dimensional vector $\vlambda = (\lambda_1, \dots, \lambda_d) \in \Lambda^d$, $\Lambda \subset \overline{\R} \coloneqq \R \cup \{\pm\infty\}$ for some general lower and upper bounds $\l_j < \u_j$ that may depend on $y$, we define $\ell^{\gamma}(x, \I_{\vlambda}(y))$ to be
\begin{equation}
    \ell^{\gamma}(x, \I_{\vlambda}(y)) = \frac{1}{d} \sum_{j \in [d]} \left[\frac{2(1 + q)}{I(\vlambda)_j} \lvert x_j - c_j \rvert - q\right]_+,
\end{equation}
where $q = \gamma/(1-\gamma)$, $\gamma \in [0, 1)$, $I(\vlambda)_j = \u_j - \l_j + 2\lambda_j$, $c_j = (\u_j + \l_j)/2$, and $[u]_+ = \max(0, u)$. First, we show that $\ell^{\gamma}(x, \I_{\vlambda}(y))$ is convex in $\vlambda$ for $\vlambda \geq 0$.
\begin{proof}
    Note that $\ell^{\gamma}(x, \I_{\vlambda}(y))$ is separable in $\vlambda$. Hence, it suffices to show that $\ell^{\gamma}(x, \I_{\vlambda}(y))$ is convex in each entry $\lambda_j$. That is, we want to show that
    \begin{equation}
        \ell^{\gamma}(x, \I_{\vlambda}(y))_j = \left[\frac{2(1 + q)}{\u_j - \l_j + 2\lambda_j} \lvert x_j - c_j \rvert - q\right]_+
    \end{equation}
    is convex in $\lambda_j$. Note that:
    \begin{itemize}
        \item The term $1/(\u_j - \l_j + 2\lambda_j)$ behaves like $1/\lambda_j$, hence it is convex for $\lambda_j \geq 0 > - (\u_j - \l_j)/2$,
        \item $C = 2(1+q)\lvert x_j - c_j \rvert$ is nonnegative, hence $C \cdot 1/(\u_j - \l_j + 2\lambda_j)$ is convex,
        \item $q$ does not depend on $\lambda_j$, hence $C \cdot 1/(\u_j - \l_j + 2\lambda_j) - q$ is still convex, and finally
        \item the positive part $[u]_+ = \max(0, u)$ is a convex function of its argument, hence $[C \cdot 1/(\u_j - \l_j + 2\lambda_j) - q]_+$ is convex.
    \end{itemize}
    We conclude that $\ell^{\gamma}(x, \I_{\vlambda}(y))_j$ is convex in each entry $j \in [d]$ for $\lambda_j \geq 0$, hence $\ell^{\gamma}(x, \I_{\vlambda}(y))$ is convex for $\vlambda \geq 0$.
\end{proof}
Note that $\ell^{\gamma}$ is an upper bound to $\ell^{01}$ by construction. One can see that $\forall \vlambda,~\ell^{\gamma}(x, \I_{\vlambda}(y)) \geq \ell^{01}(x, \I_{\vlambda}(y))$.
\begin{proof}
   We will now show that $\ell^{\gamma}(x, \I_{\vlambda}(y))_j \geq \ell^{01}(x, \I_{\vlambda}(y))_j$ entrywise. To this end, we will show that both functions attain the same value (i.e., 1) at the extremes of the intervals (i.e., $\l - \vlambda$ and $\u + \vlambda$), and use the fact that $\ell^{\gamma}(x, \I_{\vlambda}(y))_j$ is convex nonnegative. 
   
   First, note that $\forall \vlambda$, it holds that
    \begin{align}
        \ell^{\gamma}(\l - \vlambda, \I_{\vlambda}(y))_j &= \ell^{\gamma}(\u + \vlambda, \I_{\vlambda}(y))_j = \left[\frac{2(1 + q)}{I(\vlambda)_j} \cdot \bigg\lvert \frac{I(\vlambda)_j}{2} \bigg\rvert - q\right]_+ = 1,\\
         \ell^{01}(\l - \vlambda, \I_{\vlambda}(y))_j &= \ell^{01}(\u + \vlambda, \I_{\vlambda}(y))_j = 1.\\
         \intertext{Furthermore,}
         \ell^{01}(x, \I_{\vlambda}(y))_j &= 0,~\forall x_j \in [\l_j - \lambda_j, \u_j + \lambda_j]
    \end{align}
    by definition. We conclude that $\ell^{\gamma}$ upper bounds $\ell^{01}$ because $\ell^{\gamma}$ is convex, nonnegative, it intersects $\ell^{01}$ at $\l - \vlambda$ and $\u + \vlambda$ (for which both losses are equal to 1), and $\ell^{01}$ is exactly 0 between $\l_j - \lambda_j$ and $\u_j + \lambda_j$.
\end{proof}

\section{\label{supp:rcps}Risk Controlling Prediction Sets (RCPS) \cite{bates2021distribution}}
\setcounter{figure}{0}
In this section, we present the original RCPS procedure presented in \citet{bates2021distribution}. Let $\ell:~\Y \times \Y' \to \R$, $\Y' \subseteq 2^{\Y}$ be a monotonically nonincreasing loss function (see \cref{eq:monotonically_nonincreasing}) over $\X \subset \R^d$ and $\Y \subset \R^d$, and let $\{\I_{\lambda}(y)\}_{\lambda \in \Lambda}$, $\Lambda \subset \overline{\R}$ be a family of set-valued predictors $\I:~\Y \to \X'$ that satisfies the nesting property in \cref{eq:nesting}. Here, 

\cref{algo:rcps} summarizes the original conformal risk control procedure introduced in \citet{bates2021distribution} for a general bounding function $\ucb(n, \delta, \hat{R}(\lambda))$ that accepts the number of samples in a calibration set $\S_{\cal} = \{(x_i, y_i)\}_{i=1}^n$ of $n$ i.i.d. samples from an unknown distribution $\D$ over $\X \times \Y$, a failure probability $\delta \in [0, 1]$, the empirical estimate $\hat{R}(\lambda) = 1/n \cdot \sum_{i=1}^n \ell(x_i, \I_{\lambda}(y_i))$ evaluated on $\S_{\cal}$, and that returns a pointwise upper confidence bound $\hat{R}^+(\lambda)$ that satisfies
\begin{equation}
    \P[R(\lambda) \leq \hat{R}^+(\lambda)] \geq 1 - \delta
\end{equation}
as presented in \cref{eq:ucb}. For example, for losses bounded above by 1, Hoeffding's inequality \cite{hoeffding1994probability} yields
\begin{equation}
    \hat{R}^+(\lambda) = \text{UCB}(n, \delta, \hat{R}(\lambda)) = \hat{R}(\lambda) + \sqrt{\frac{1}{2n}\log\left(\frac{1}{\delta}\right)}.
\end{equation}
In practice, tighter alternatives exist (see \citet{bates2021distribution}, Section~3.1 for a thorough discussion).

\begin{algorithm}[h]
   \caption{\label{algo:rcps}$(\e, \delta)$-RCPS (see \cite{angelopoulos2022image}, Algorithm~2)}
\begin{algorithmic}[1]
    \STATE {\bfseries Input:} risk level $\e \geq 0$, failure probability $\delta \in [0, 1]$, calibration set $\S_{\cal} = \{(x_i, y_i)\}_{i=1}^n$ of $n$ i.i.d. samples, monotonically nonincreasing loss function $\ell$, family of nested set-valued predictors $\{\I_\lambda\}_{\lambda \in \Lambda}$, $\Lambda \subset \overline{\R}$, initial (large) value $\lambda_{\max}$, stepsize $\d{\lambda} > 0$.
    \STATE $\lambda \gets \lambda_{\max}$
    \STATE $\hat{R}^+(\lambda) \gets 0$
    \WHILE{$\hat{R}^+(\lambda) \leq \e$}
        \STATE $\lambda \gets \lambda - \d{\lambda}$
        \STATE $\hat{R}(\lambda) \gets 1/n \cdot \sum_{(x_i, y_i) \in \S_{\cal}} \ell(x_i, \I_{\lambda}(y_i))$
        \STATE $\hat{R}^+(\lambda) \gets \text{UCB}(n, \delta, \hat{R}(\lambda))$
    \ENDWHILE
    \STATE $\hat{\lambda} \gets \lambda + \d{\lambda}$
    \STATE {\bfseries return} $\hat{\lambda}$
\end{algorithmic}
\end{algorithm}

\section{\label{supp:ucb_01}Computing the UCB}
In this section, we include a detailed analytical example on how one can compute the UCB $\hat{R}^{01+}(\vlambda + \veta)$ by means of concentration inequalities. Recall that, for a family of nested set predictors $\{\I_{\vlambda}(y)\}_{\vlambda \in \Lambda^d}$ indexed by a \emph{vector} $\vlambda = (\lambda_1, \dots, \lambda_d)$, the $01$ loss $\ell^{01}(x, \I_{\vlambda}(y))$ counts the number of pixels in $x$ that fall outside of their respective intervals in $\I_{\vlambda}(y)$. For example, choosing $\I_{\vlambda}(y)_j = [\l_j - \lambda_j, \u_j + \lambda_j]$ as proposed in \cref{eq:scalar_additive_interval}, given a vector $\veta = (\eta_1, \dots, \eta_d)$ yields
\begin{equation}
    \ell^{01}(x, \I_{\vlambda + \veta}(y)) = \frac{1}{d} \sum_{j \in [d]} \1\left[x_j \notin [\l_j - (\lambda_j + \eta_j), \u_j + (\lambda_j + \eta_j)]\right].
\end{equation}
Then, $R^{01}(\vlambda + \veta) = \E[\ell^{01}(x, \I_{\vlambda + \veta}(y))]$ denotes the risk of $\I_{\vlambda + \veta}$ and its empirical estimate over a calibration set $\S_{\cal} = \{(x_i, y_i)\}_{i=1}^n$ of $n$ i.i.d. samples from an unknown distribution $\D$ over $\X \times \Y$ is
\begin{equation}
    \hat{R}^{01}(\vlambda + \veta) = \frac{1}{n} \sum_{i \in [n]} \ell^{01}(x_i, \I_{\vlambda + \veta}(y_i)).
\end{equation}
In turn, for a user-specified failure probability $\delta \in [0, 1]$, the UCB $R^{01+}(\vlambda + \veta)$ can be obtained by means of common concentration inequalities. For example, Hoeffding's inequality implies
\begin{equation}
    \hat{R}^{01+}(\vlambda + \veta) = \hat{R}^{01}(\vlambda + \veta) + \sqrt{\frac{1}{2n}\log\left(\frac{1}{\delta}\right)}.
\end{equation}

\section{\label{supp:experimental_details}Experimental Details}
\setcounter{figure}{0}
In this section, we include further experimental details. All experiments were performed on a private cluster with 8 NVIDIA RTX A5000 with 24 GB of memory.

\subsection{Datasets}

\begin{figure*}[t]
    \centering
    \subcaptionbox{\label{supp:example_celeba}CelebA dataset.}{\includegraphics[width=0.48\linewidth]{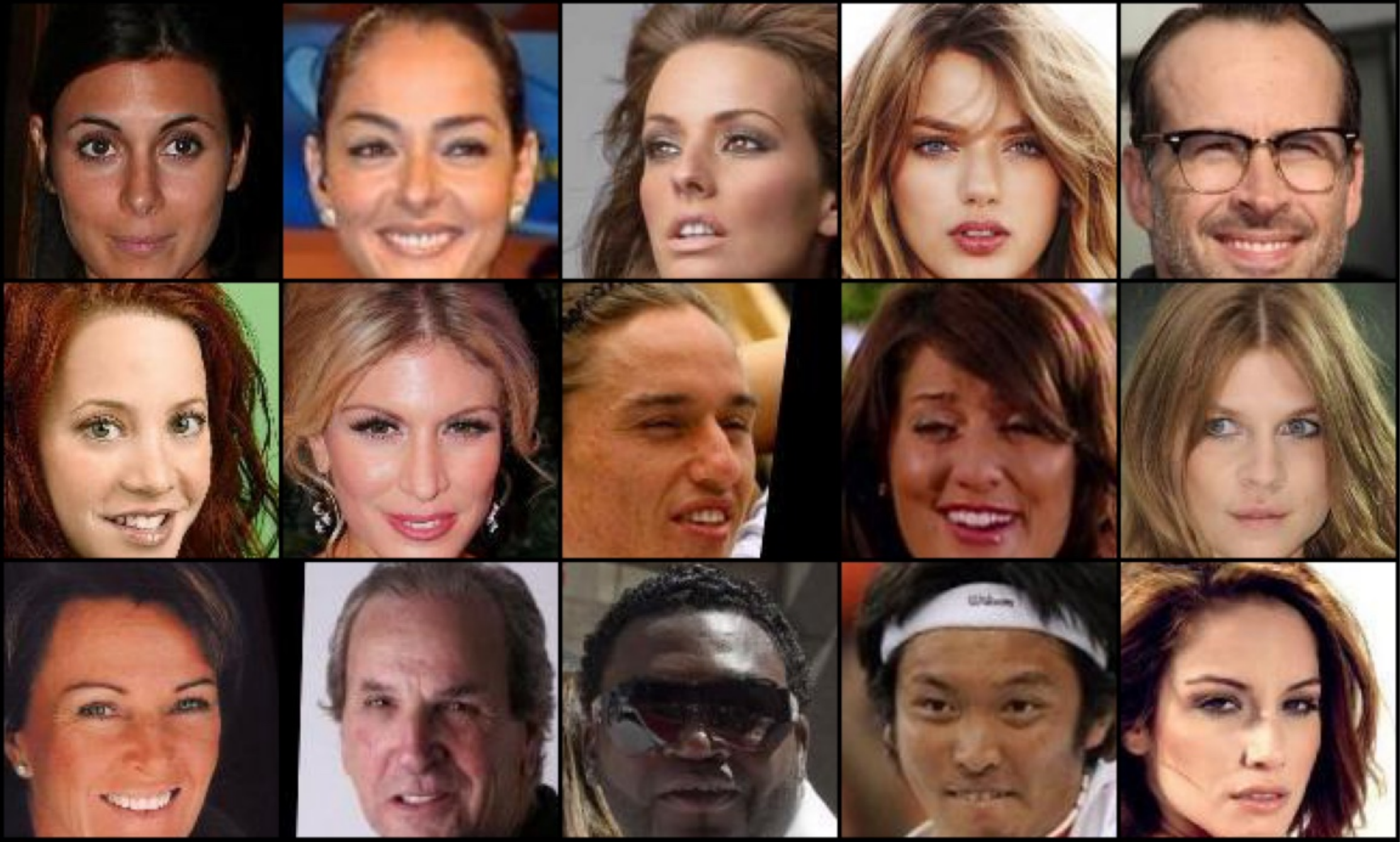}}
    \subcaptionbox{\label{supp:example_abdomen}AbdomenCT-1K dataset.}{\includegraphics[width=0.48\linewidth]{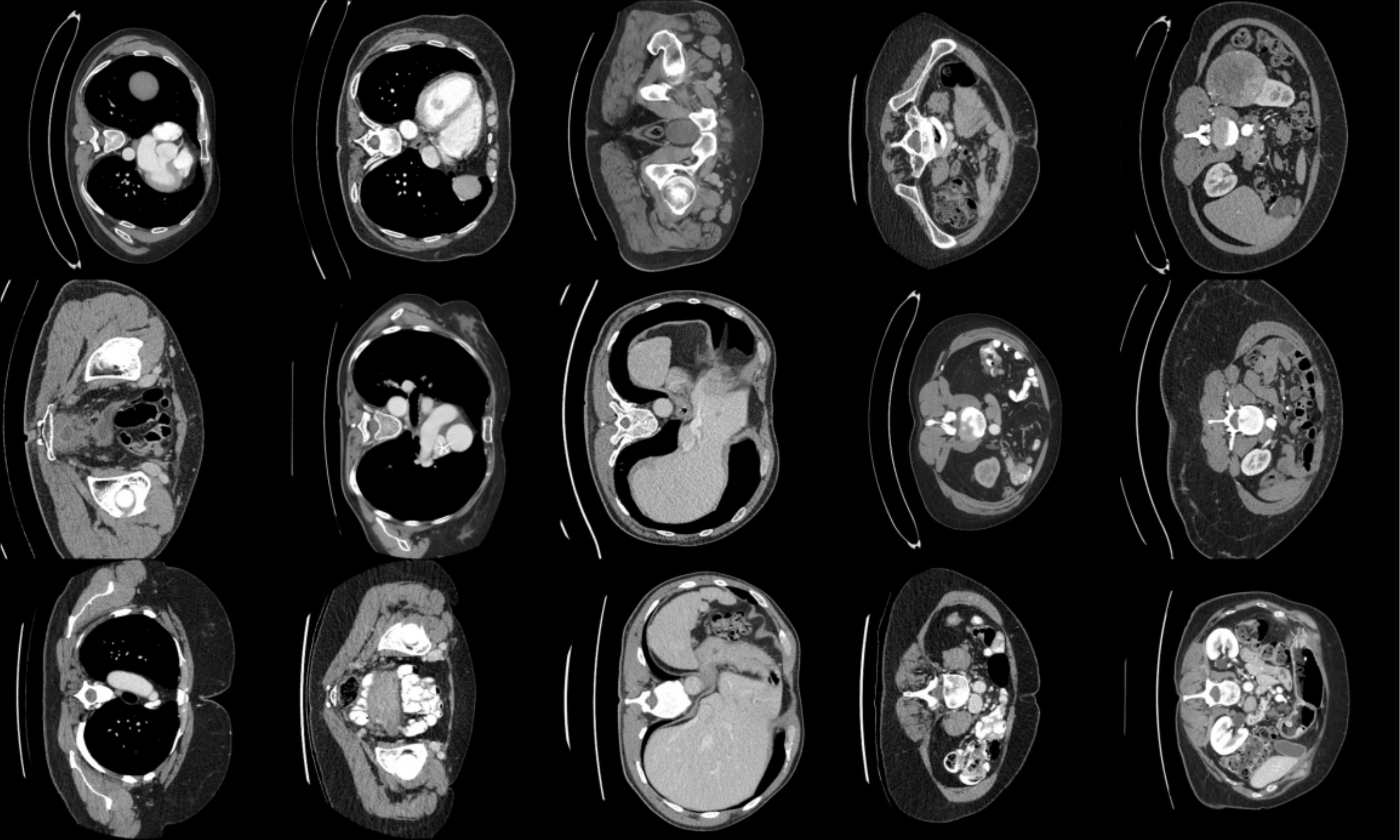}}
    \caption{\label{supp:example}Example images.}
\end{figure*}

\subsubsection{CelebA dataset}
The CelebA dataset \cite{liu2018large} (available at \url{http://mmlab.ie.cuhk.edu.hk/projects/CelebA.html}) contains more than $200 \times 10^3$, $178 \times 218$ pixel images of celebrity faces with several landmark locations and binary attributes (e.g., eyeglasses, bangs, smiling). In this work, we center-crop all images to $128 \times 128$ pixels and normalize them between $[-1, 1]$.

\subsubsection{AbdomenCT-1K dataset}
The AbdomenCT-1K dataset \cite{ma2021abdomenct} (available at \url{https://github.com/JunMa11/AbdomenCT-1K}) comprises more than 1000 abdominal CT scans (for a total of more than $200 \times 10^3$, $512 \times 512$ pixel individual images) aggregated from 6 existing datasets. Scans are provided in NIfTI format, so we first convert them to their Hounsfield unit (HU) values and subsequently window them with the standard abdomen setting (WW = 400, WL = 40) such that pixels intensities are normalized in $[0, 1]$ and they represent the same tissue across images.

\cref{supp:example} shows some example images for both datasets.

\subsection{Models}
Recall that in this work, we train both:
\begin{itemize}
    \item A score network $s(\tilde{x}, t) \approx \grad_x \log p_t(\tilde{x})$ to sample from $p(x | y)$ as introduced in \cref{sec:conditional_sampling}, and
    \item a modified time-conditional image regressor $f:~\Y \times \R \to \X^3$ such that $f(y, t) = (\q_{\alpha/2}, \hat{x}, \q_{\alpha/2})$, where $\hat{x} \approx \E[x \mid y]$ and $\q_{\alpha/2},\q_{1 - \alpha/2}$ are the entrywise $\alpha/2$ and $1 - \alpha/2$ quantile regressors of $x$ \cite{koenker1978regression,romano2019conformalized,angelopoulos2022image}, respectively.
\end{itemize}
Both models are implemented wit the same U-net-like \cite{ronneberger2015u} backbone: NCSN++, which was introduced by \citet{song2020score} (code is available at \url{https://github.com/yang-song/score_sde}). We use the original NCSN++ configurations presented in \citet{song2020score} for the CelebA dataset and, for the AbdomenCT-1K dataset, for the FFHQ dataset (available at \url{https://github.com/NVlabs/ffhq-dataset}) given the larger image size of $512 \times 512$. For the image regressor $f$, we use the original implementation of the quantile regression head used in \citet{angelopoulos2022image} (available at \url{https://github.com/aangelopoulos/im2im-uq}) on top of the NCSN++ backbone. This allows us to maintain a time-conditional backbone and extend the original image regressor presented in \citet{angelopoulos2022image} to all noise levels as the score network.

\subsection{Training}

\subsubsection{Data Augmentation}
For both datasets, we augment the training data by means of random horizontal and vertical flips, and random rotations between $[-\pi/2, \pi/2]$ degrees.

\subsubsection{Forward SDE}

\begin{figure*}[t]
    \centering
    \subcaptionbox{\label{supp:example_forward_celeba}CelebA dataset.}{\includegraphics[width=0.48\linewidth]{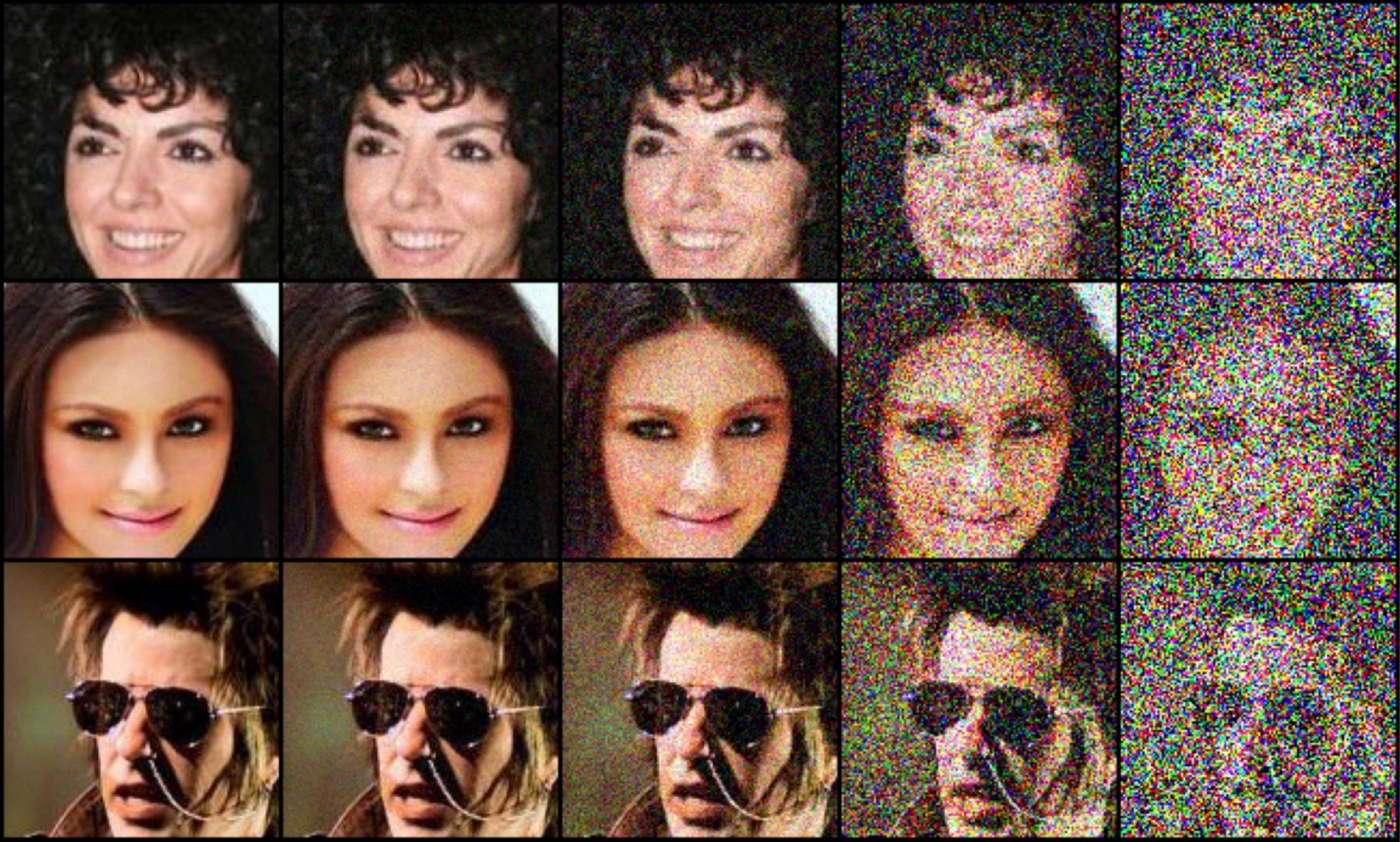}}
    \subcaptionbox{\label{supp:example_forward_abdomen}AbdomenCT-1K dataset.}{\includegraphics[width=0.48\linewidth]{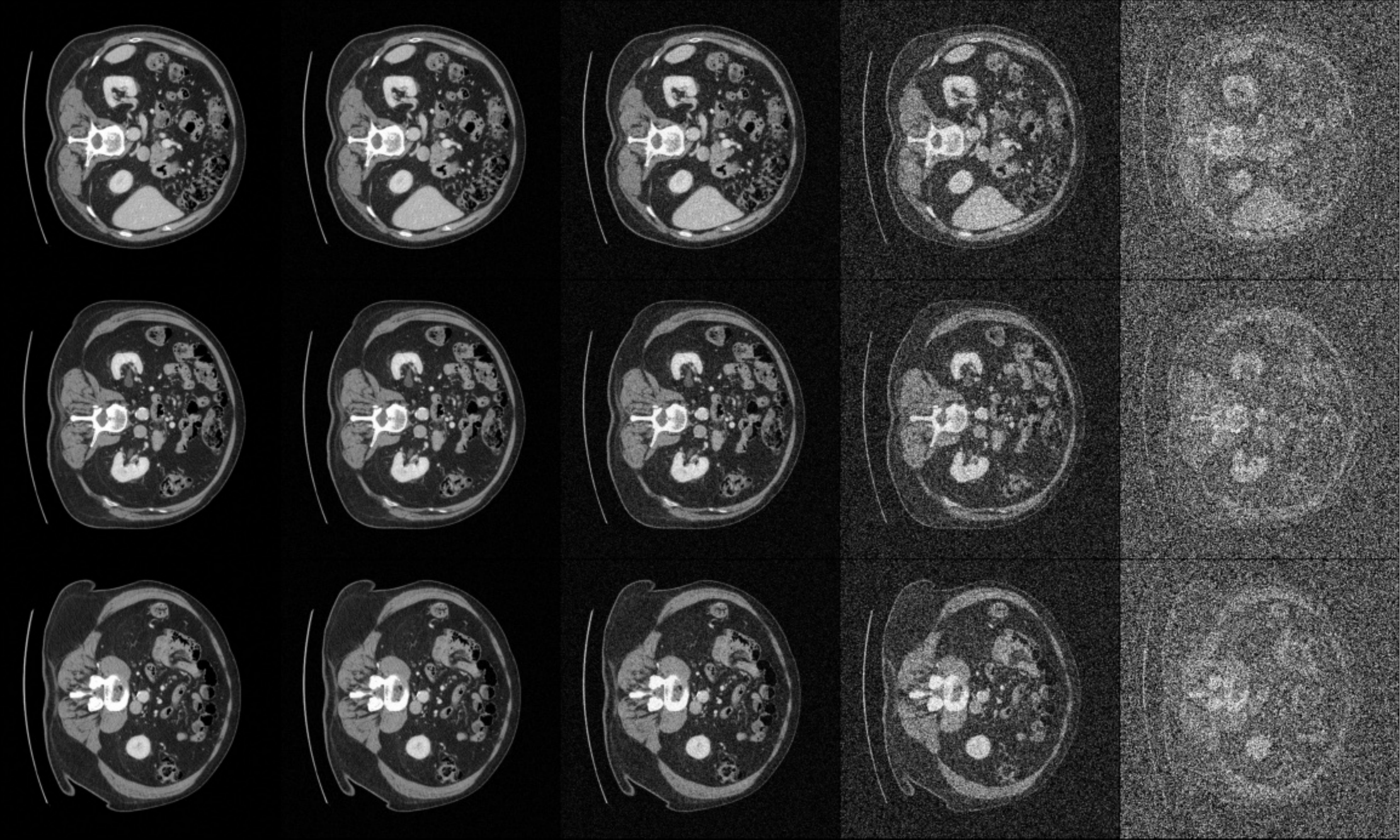}}
    \caption{\label{supp:example_forward}Example of perturbed images via the forward SDE. The final level of noise is $\sigma^2 = 1$.}
\end{figure*}

Recall that in this work, we are interested in solving the classical denoising problem where $y = x + v_0$ is a noisy observation of a ground truth image $x$ perturbed with random Gaussian noise with known variance $\sigma_0^2$. As done by previous works \cite{song2019generative,song2020improved,song2021solving,kawar2021snips}, we model the observation process with a \emph{variance exploding} (VE) forward-time SDE
\begin{equation}
    \label{eq:forward_sde}
    \d{x} = \sqrt{\frac{\d{[\sigma^2(t)]}}{\d{t}}}\,\d{w}    \quad~\text{where}~\quad    \sigma(t) = \sigma_{\min} \left(\frac{\sigma_{\max}}{\sigma_{\min}}\right)^t,~t \in [0, 1]
\end{equation}
such that $\sigma(0) = \sigma_{\min}$ and $\sigma(1) = \sigma_{\max}$. In particular, we set $\sigma_{\min} = 0$ and $\sigma_{\max} = 90$ for the CelebA dataset and $\sigma_{\max} = 1$ for the AbdomenCT-1K dataset. It has been shown \cite{song2020score} that the above forward-time SDE is equivalent to the following discrete Markov chain
\begin{equation}
    x_{t} = x_{t-1} + \sqrt{\sigma^2_{t} - \sigma^2_{t-1}} z,~z \sim \N(0, \Id),    \quad~\text{for}~\quad  t = 1, \dots, N
\end{equation}
and $\{\sigma_i\}_{i=0}^N$ noise levels. That is, $x_t = x + z_t$, where $z_t \sim (0, \sigma_t^2)$. \cref{supp:example_forward} shows some example images from both datasets perturbed via the forward SDE described in this section.

\subsubsection{Denoising Score-matching}
Here, we briefly describe the loss function used to train the time-conditional score network $s(\tilde{x}, t)$. Denote $\theta \in \Theta$ the parametrization of $s$, then, following \citet{song2019generative,song2020score}, we have $s(\tilde{x}, t) = s_{\theta^*}(\tilde{x}, t)$, where
\begin{align}
    \theta^*    &= \argmin_{\theta \in \Theta} \E_{t \sim U(0, 1)}\left[\xi(t) \E_{x \sim p(x),~x(t) | x}\left[\|s_{\theta}(x(t), t) - \grad_{x} \log p_t(x(t) | x)\|^2\right]\right]\\
                &= \argmin_{\theta \in \Theta} \E_{t \sim U(0, 1)}\left[\xi(t) \E_{x \sim p(x),~x(t) | x}\left[\bigg\|s_{\theta}(x(t), t) + \frac{x(t) - x}{\sigma(t)}\bigg\|^2\right]\right],
\end{align}
where $\xi(t) \propto \sigma^2(t)$ and $U(0, 1)$ is the uniform distribution over $[0, 1]$.

\subsubsection{Quantile Regression}

\begin{figure*}[t]
    \centering
    \subcaptionbox{\label{supp:example_quantile_celeba}CelebA dataset.}{\includegraphics[width=0.48\linewidth]{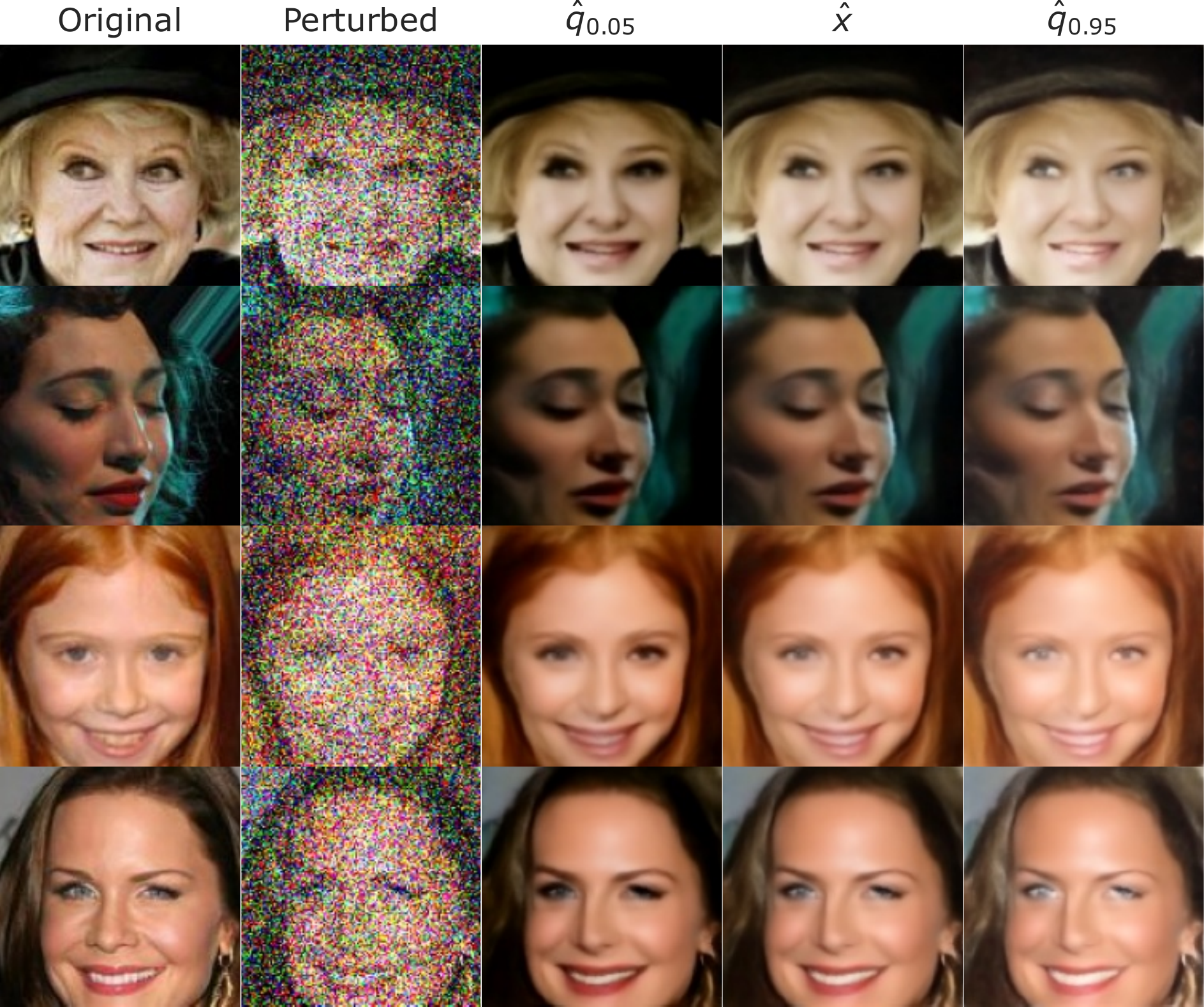}}
    \subcaptionbox{\label{supp:example_quantile_abdomen}AbdomenCT-1K dataset.}{\includegraphics[width=0.48\linewidth]{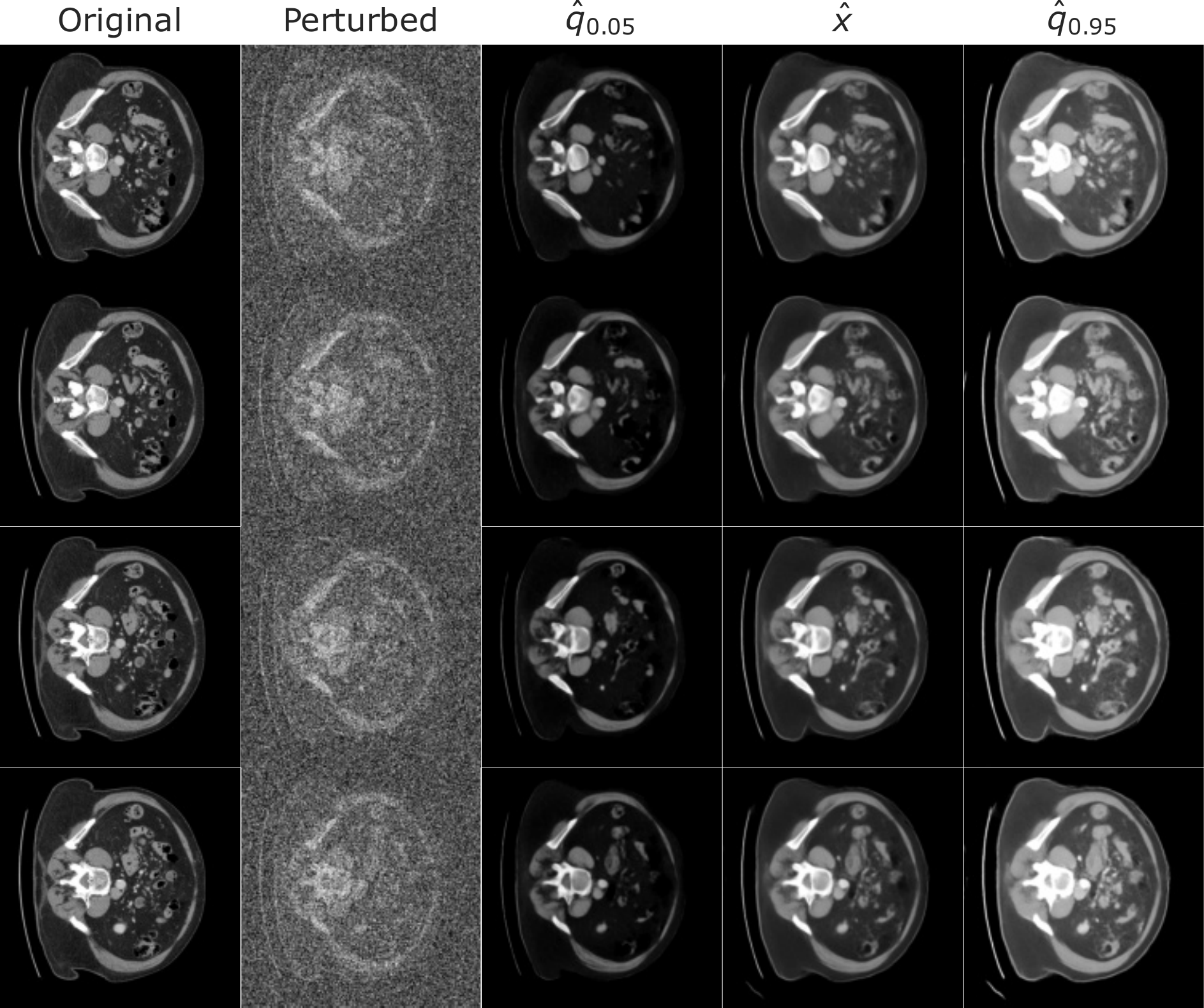}}
    \caption{\label{supp:example_quantile}Example of results with the modified image regressor.}
\end{figure*}

Similarly to above, we briefly describe the loss function used to train the modified time-conditional image regressor $f$. Denote $\theta \in \Theta$ the parametrization of $f$, and recall that for a desired $\alpha$ quantile and its respective quantile regressor $\q_{\alpha}$, the \emph{quantile loss function} \cite{koenker1978regression,romano2019conformalized,angelopoulos2022image} $\ell^{\alpha}(x, \q_{\alpha})$ is 
\begin{equation}
    \ell^{\alpha}(x, \q_{\alpha}) = \alpha (x - \q_{\alpha}) \cdot \1[x > \q_{\alpha}] + (1 - \alpha) (\q_{\alpha} - x) \cdot \1[x \leq \q_{\alpha}].
\end{equation}
For the image regressor $f$, we set $\alpha = 0.10$ and use the original implementation of the quantile regression head from \citet{angelopoulos2022image} (available at \url{https://github.com/aangelopoulos/im2im-uq}) which minimizes the multi-loss objective
\begin{equation}
    \mathcal{L}(x, f(y, t)) = \ell^{\alpha/2}(x, \q_{\alpha/2}) + (x - \hat{x})^2 + \ell^{1 - \alpha/2}(x, \q_{1 - \alpha/2})
\end{equation}
on top of the NCSN++ backbone. This allows us to maintain a time-conditional backbone and extend the original image regressor presented in \citet{angelopoulos2022image} to all noise levels as the score network. Then, $f(y, t) = f_{\theta^*}(y, t)$ where
\begin{equation}
    \theta^* = \argmin_{\theta \in \Theta} \E_{t \sim U(0,1)}\left[\mathcal{L}(x, f(x(t), t)\right].
\end{equation}

\subsection{Sampling}

\begin{figure*}[h]
    \centering
    \subcaptionbox{\label{supp:example_sample_celeba}CelebA dataset.}{\includegraphics[width=0.48\linewidth]{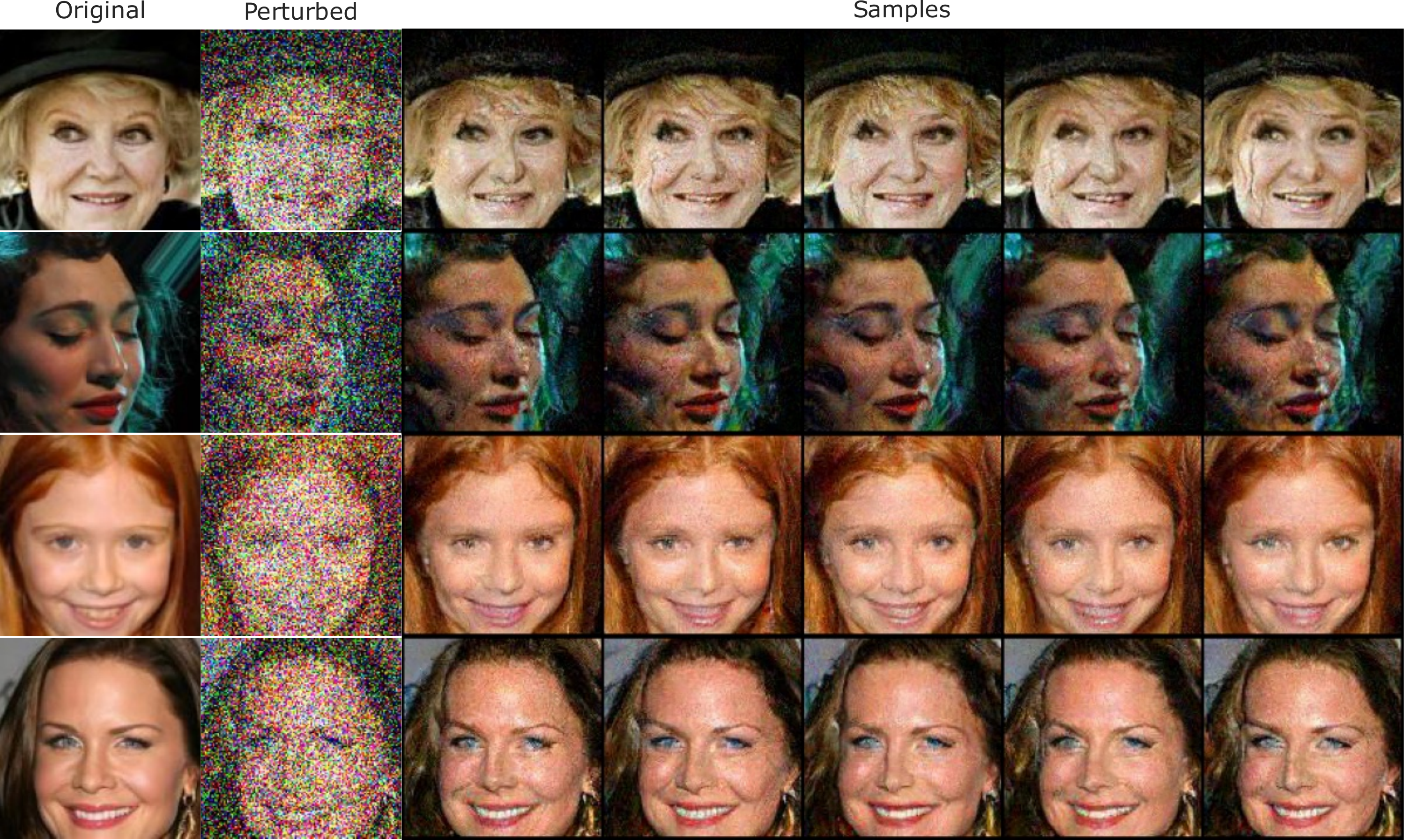}}
    \subcaptionbox{\label{supp:example_sample_abdomen}AbdomenCT-1K dataset.}{\includegraphics[width=0.48\linewidth]{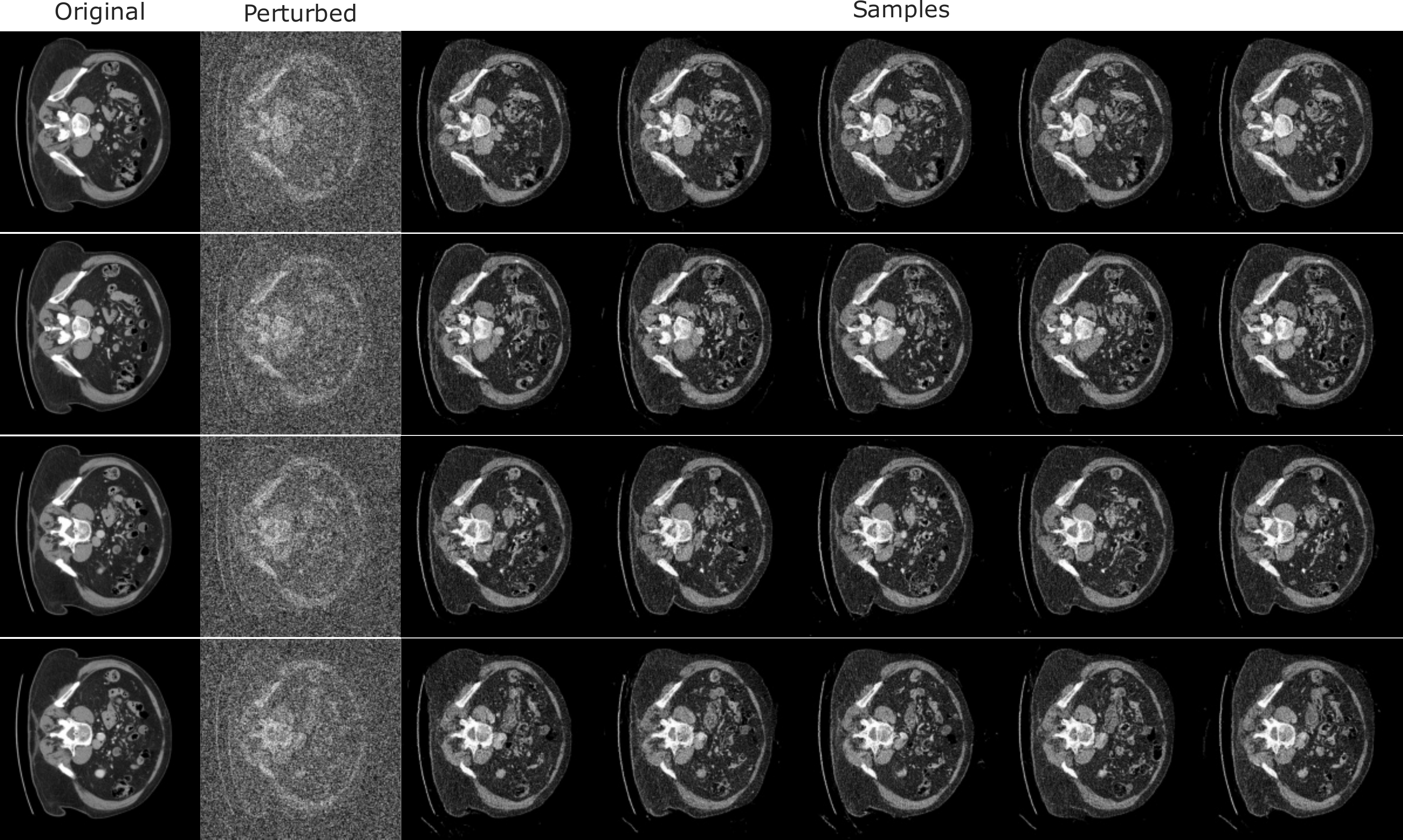}}
    \caption{\label{supp:example_sample}Example of images sampled via \cref{algo:reverse_sde}.}
\end{figure*}

Here, we briefly describe the sampling procedure used in this work to solve the conditional reserve-time SDE
\begin{equation}
    \d{x} = [h(x, t) - g(t)^2 \grad_x \log p_t(x | y)]\,\d{t} + g(t)\,\d{\bar{w}}
\end{equation}
where $y$ is the initial noisy observation with known noise level $\sigma_0^2$, and $g(t) = \sqrt{\d{[\sigma^2(t)]}/\d{t}},~h(x,t) = 0$ as in \cref{eq:forward_sde}. Although several different discretization schemes exist \cite{song2020score}, we use the classical Euler-Maruyama discretization \cite{karatzas1991brownian} since the contribution of this work is not focused on improving existing diffusion models. Recall that for a general It\^{o} process
\begin{equation}
    \d{x} = h(x,t)\,\d{t} + g(t)\,\d{w},
\end{equation}
its Euler-Maruyama discretization with step $\Delta{t}$ is
\begin{equation}
    x_{t+1} = x_t + h(x,t)\,\Delta{t} + g(t)\sqrt{\Delta{t}}\,z,~z \sim \N(0, \Id).
\end{equation}
Furthermore, under the forward SDE described in \cref{eq:forward_sde}, and as shown by previous works \cite{kawar2021stochastic,kawar2021snips,kadkhodaie2021stochastic}, it holds that 
\begin{equation}
    \grad_x \log p_t(x(t) | y) = \grad_x \log p_t(x(t)) + \frac{y - x(t)}{\sigma^2_0 - \sigma^2(t)}.
\end{equation}
We conclude that the reverse-time Euler-Maruyama discretization of the conditional SDE above is
\begin{equation}
    x_{t-1} = x_t + g^2_{t}\left[\grad_x \log p_t(x_t) + \frac{y - x_t}{\sigma^2_0 - \sigma^2_t}\right]\,\Delta{t} + g_t\sqrt{\Delta{t}}\,z,~z \sim \N(0, \Id).
\end{equation}
\cref{algo:reverse_sde} implements the sampling procedure for a general score network $s(\tilde{x}, t)$.

\begin{algorithm}[h]
   \caption{\label{algo:reverse_sde}Denoising Reverse-time SDE}
\begin{algorithmic}[1]
    \STATE {\bfseries Input:} observation $y$, initial noise level $\sigma_0$, $\sigma_{\max},\sigma_{\min}$, number of steps $N$
    \STATE $t_0 \gets (\log \sigma_0 - \log \sigma_{\min})/(\log \sigma_{\max} - \log \sigma_{\min})$
    \STATE $\Delta{t} \gets 1/N$
    \STATE $n \gets \lfloor t_0/\Delta{t} \rfloor$
    \STATE $x_n \gets y$
    \FOR{$i = n, \dots, 1$}
        \STATE Draw $z_i \sim \N(0, \Id)$
        \STATE $t_i \gets i \Delta t$
        \STATE $\sigma_i \gets \sigma_{\min} \cdot (\sigma_{\max} / \sigma_{\min})^{t_i}$
        \STATE $g_i \gets \sigma_i \cdot \sqrt{2 \log(\sigma_{\max}/\sigma_{\min})}$
        \STATE $x_{i-1} \gets x_i + g^2_i[s(x_i, t_i) + (y - x_i)/(\sigma_0^2 - \sigma^2_i)]\,\Delta{t} + g_i\sqrt{\Delta{t}}\,z_i$
    \ENDFOR
    \STATE {\bfseries return} $x_0$
\end{algorithmic}
\end{algorithm}

\section{\label{supp:comparision_conffusion}Comparison with Con\textbf{\textit{ffusion}}}
In this section, we discuss the differences between the Con$ffusion$ framework proposed by \citet{horwitz2022conffusion} and the contributions of this paper. Although similar in spirit and motivation, the two are fundamentally different and address separate and distinct questions. Con$ffusion$ deploys ideas of quantile regression \cite{koenker1978regression} to fine-tune an existing score network and obtain an image regressor equipped with heuristic uncertainty intervals. Such intervals can then be conformalized to provide risk control. On the other hand, our $K$-RCPS is a novel high-dimensional calibration procedure for any stochastic sampler and any notion of uncertainty (i.e., it is agnostic of the notion of uncertainty), including diffusion models and quantile regression. In this paper, we present $K$-RCPS for diffusion models given their remarkable performance in solving inverse problems via conditional sampling.

In order to contextualize the choices of parametrization for the families of set predictors in \cref{eq:set_conffusion_multiplicative,eq:set_conffusion_additive}, we highlight a gap between the presentation of Con$ffusion$ in the arXiv paper available at \url{https://arxiv.org/abs/2211.09795v1} and the official code release in \url{https://github.com/eliahuhorwitz/Conffusion}. In particular Section~3.2 of \citet{horwitz2022conffusion} introduces the following family of set predictors:
\begin{equation}
    \I^{~\text{arXiv}}_{\vlambda,\text{Conffusion}}(y)_j = [\lambda_j(\q_{\alpha/2})_j, \lambda_j(\q_{1 - \alpha/2})_j],
\end{equation}
which does not satisfy the nesting property in \cref{eq:nesting}. It is easy to see that---without any further assumptions---as $\lambda_j$ increases, $\I^{~\text{arXiv}}_{\vlambda,\text{Conffusion}}(y)_j$ does not cover the interval $[0,1]$ and the $01$ loss in \cref{eq:01_loss} is not entrywise monotonically non-increasing. This renders the RCPS procedure (and most calibration procedures) not applicable as presented in the arXiv paper. While this family of set predictors is thus not applicable, we resorted to the authors' GitHub release where such parametrization is different (see \url{https://github.com/eliahuhorwitz/Conffusion/blob/fffe5c946219cf9dead1a1c921a131111e31214e/inpainting_n_conffusion/core/calibration_masked.py#L28}). More precisely, the GitHub implementation reflects the multiplicative parametrization presented in \cref{eq:set_conffusion_multiplicative}, which does indeed satisfy the entrywise nesting property. Finally, we additionally propose the additive parametrization in \cref{eq:set_conffusion_additive} to showcase how $K$-RCPS is agnostic of the notion of uncertainty and it can be used in conjunction with Con$ffusion$.

\newpage
\clearpage

\section{\label{supp:errata}Erratum of \cref{thm:multidimensional_risk_control}}
In this section, we address the differences between the version of \cref{thm:multidimensional_risk_control} presented in this manuscript and the published paper available at \url{https://proceedings.mlr.press/v202/teneggi23a.html}, which contains a small typo. We include the two theorems side-by-side with differences highlighted.

\vspace{1em}
\noindent
\begin{minipage}[t]{0.48\textwidth}
\begin{theorem*}[This manuscript]
    Let $\ell:~\X \times \X' \to \R$, $\X' \subseteq 2^{\X}$, $\X \subset \R^d$ be an entrywise monotonically nonincreasing function and let $\{\I_{\vlambda}(y) = [\l_j - \lambda_j, \u_j + \lambda_j]\}_{\vlambda \in \Lambda^d}$ be a family of set-valued predictors $\I:~\Y \to \X'$, $\Y \subset \R^d$ indexed by $\vlambda \in \Lambda^d$, $\Lambda \subset \overline{\R}$, for some lower and upper bounds $\l_j < \u_j$ that may depend on $y$. {\color{ForestGreen}Given $\tLambda = \widetilde{\vlambda} + \omega\veta,~\omega \in \Lambda$, for fixed $\widetilde{\vlambda} \in \R^d$ and $\veta \in \R^d,~\veta \geq 0$}, if
    \begin{align*}
        \hat{\vlambda}  &= \argmin_{{\color{ForestGreen}\vlambda \in \tLambda }}~\sum_{j \in [d]} \lambda_j   \quad\st\quad\hat{R}^+(\vlambda + \beta\veta) < \e
    \end{align*}
    $\forall \beta \geq 0$, then $\I_{\hat{\vlambda}}(y)$ is an $(\e, \delta)$-RCPS and $\hat{\vlambda}$ minimizes the mean interval length {\color{ForestGreen}over $\tLambda$}.
\end{theorem*}
\end{minipage}
\hfill
\vrule
\hfill
\begin{minipage}[t]{0.48\textwidth}
\begin{theorem*}[Published paper]
    Let $\ell:~\X \times \X' \to \R$, $\X' \subseteq 2^{\X}$, $\X \subset \R^d$ be an entrywise monotonically nonincreasing function and let $\{\I_{\vlambda}(y) = [\l_j - \lambda_j, \u_j + \lambda_j]\}_{\vlambda \in \Lambda^d}$ be a family of set-valued predictors $\I:~\Y \to \X'$, $\Y \subset \R^d$ indexed by $\vlambda \in \Lambda^d$, $\Lambda \subset \overline{\R}$, for some lower and upper bounds $\l_j < \u_j$ that may depend on $y$. For a fixed vector $\veta \in \R^d,~\veta \geq 0$, if
    \begin{align*}
        \hat{\vlambda}  &= \argmin_{\vlambda \in \Lambda^d}~\sum_{j \in [d]} \lambda_j   \quad\st\quad\hat{R}^+(\vlambda + \beta\veta) < \e
    \end{align*}
    $\forall \beta \geq 0$, then $\I_{\hat{\vlambda}}(y)$ is an $(\e, \delta)$-RCPS, and $\hat{\vlambda}$ minimizes the mean interval length.
\end{theorem*}
\end{minipage}

\paragraph{Reason for the differences}
The domain of the optimization problem in the published paper, i.e. $\vlambda \in \Lambda^d$, does not reflect the \emph{pointwise} coverage guarantee provided by $\hat{R}^+$. That is, for each \emph{fixed} $\vlambda \in \Lambda^d$, $\P[R(\vlambda) \leq \hat{R}^+(\vlambda)] \geq 1 - \delta$. To solve the optimization problem over $\vlambda \in \Lambda^d$, $\hat{R}^+$ should provide \emph{uniform} coverage, i.e. $\forall \vlambda \in \Lambda^d,~\P[R(\vlambda) \leq \hat{R}^+(\vlambda)] \geq 1 - \delta$.

\paragraph{Effect on results}
We remark that the changes in \cref{thm:multidimensional_risk_control} do not affect the $K$-RCPS procedure nor the results presented in the published version of the paper.

\newpage
\clearpage

\section{Figures}
\setcounter{figure}{0}
This section contains supplementary figures.

\begin{figure*}[h!]
    \centering
    \subcaptionbox{\label{fig:risk_control_celeba}CelebA dataset.}{\includegraphics[width=0.48\linewidth]{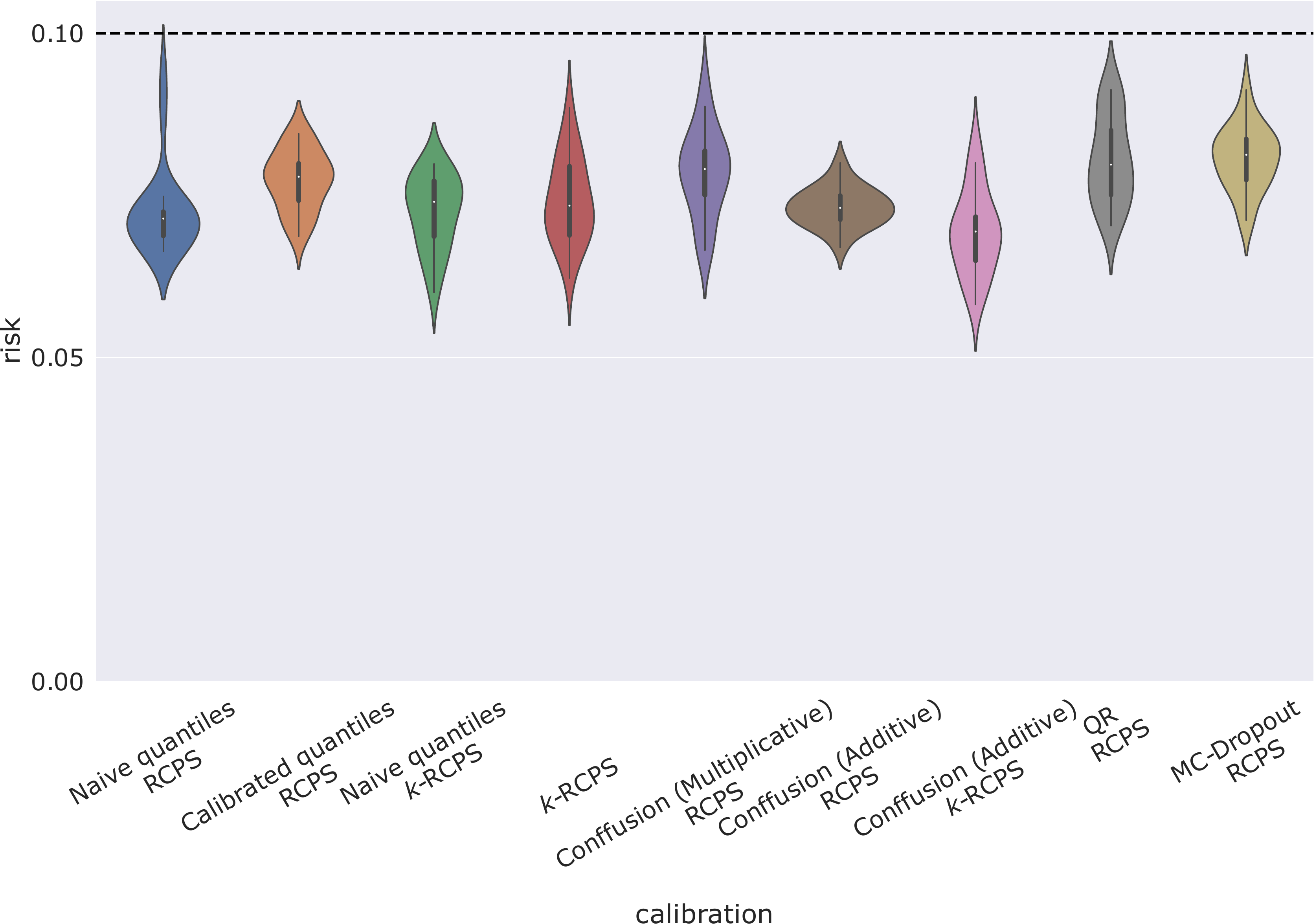}}
    \subcaptionbox{\label{fig:risk_control_abdomen}AbdomenCT-1K dataset.}{\includegraphics[width=0.48\linewidth]{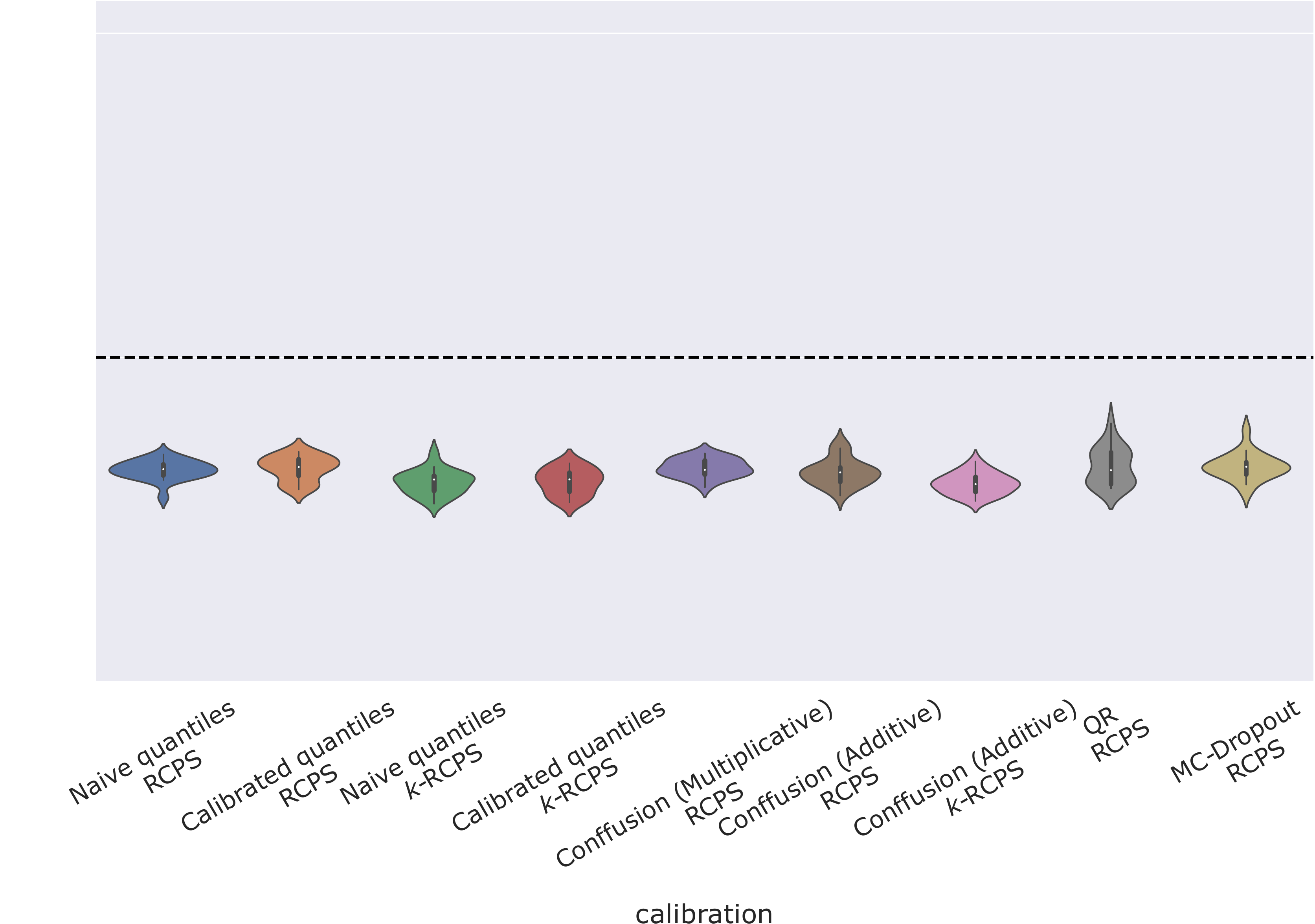}}
    \vspace{-5pt}
    \caption{\label{fig:risk_control}Empirical estimates of risk over 20 random draws of $\S_{\cal}$. All combinations of notion of uncertainty and calibration procedure successfully control risk at level $\e = 0.10, 0.05$ for the CelebA and AbdomenCT-1K dataset, respectively, with probability at least $1 - \delta,~\delta = 0.10$.}
    \vspace{-10pt}
\end{figure*}

\end{document}